\documentclass[sigconf]{acmart}

\usepackage{multirow}
\usepackage[ruled,linesnumbered]{algorithm2e}
\usepackage{amsmath}
\usepackage[flushleft]{threeparttable}
\usepackage{balance}

\AtBeginDocument{%
  \providecommand\BibTeX{{%
    \normalfont B\kern-0.5em{\scshape i\kern-0.25em b}\kern-0.8em\TeX}}}

\setcopyright{acmcopyright}
\copyrightyear{2018}
\acmYear{2018}
\acmDOI{10.1145/1122445.1122456}

\acmConference[Woodstock '18]{Woodstock '18: ACM Symposium on Neural
  Gaze Detection}{June 03--05, 2018}{Woodstock, NY}
\acmBooktitle{Woodstock '18: ACM Symposium on Neural Gaze Detection,
  June 03--05, 2018, Woodstock, NY}
\acmPrice{15.00}
\acmISBN{978-1-4503-XXXX-X/18/06}

\acmSubmissionID{fp5182259}

\begin{document}

\title[Multiplex Bipartite Network Embedding using Dual Hypergraph Convolutional Networks]{Multiplex Bipartite Network Embedding using\\ Dual Hypergraph Convolutional Networks}

\fancyhead{}


\author{Hansheng Xue}
\affiliation{
  \institution{The Australian National University}
  }
\email{hansheng.xue@anu.edu.au}

\author{Luwei Yang}
\affiliation{
  \institution{Alibaba Group}
  }
\email{luwei.ylw@alibaba-inc.com}

\author{Vaibhav Rajan}
\affiliation{
  \institution{National University of Singapore}
  }
\email{vaibhav.rajan@nus.edu.sg}

\author{Wen Jiang}
\affiliation{
  \institution{Alibaba Group}
  }
\email{wen.jiangw@alibaba-inc.com}

\author{Yi Wei}
\affiliation{
  \institution{Alibaba Group}
  }
\email{yi.weiy@alibaba-inc.com}

\author{Yu Lin}
\authornote{Corresponding author.}
\affiliation{
  \institution{The Australian National University}
  }
\email{yu.lin@anu.edu.au}

\renewcommand{\shortauthors}{Xue and Lin, et al.}


\begin{abstract}
A bipartite network is a graph structure where nodes are from two distinct domains and only inter-domain interactions exist as edges. 
A large number of network embedding methods exist to learn vectorial node representations from general graphs with both homogeneous and heterogeneous node and edge types, including some that can specifically model the distinct properties of 
bipartite networks.
However, these methods are inadequate to model multiplex bipartite networks (e.g., in e-commerce), that have multiple types of interactions (e.g., click, inquiry, and buy) and node attributes.
Most real-world multiplex bipartite networks are also sparse and have imbalanced node distributions that are challenging to model.
In this paper, we develop an unsupervised \textbf{Dual} \textbf{H}yper\textbf{G}raph \textbf{C}onvolutional \textbf{N}etwork (\textbf{DualHGCN}) model that scalably 
transforms the multiplex bipartite network 
into two sets of homogeneous hypergraphs and 
uses spectral hypergraph convolutional operators, 
along with intra- and inter-message passing strategies to promote information exchange within and across domains,
to learn effective node embeddings.
We benchmark DualHGCN using four real-world datasets on link prediction and node classification tasks. 
Our extensive experiments demonstrate that DualHGCN significantly outperforms  state-of-the-art methods, and is robust to varying sparsity levels and imbalanced node distributions. 
\end{abstract}





\maketitle

\section{Introduction}
Network representation learning aims to learn low-dimensional real-valued features of its nodes, also called embeddings, to
capture the global structural information of the network~\cite{Cui2019ASO,Cai2018ACS}. 
Such vectorial representations enable their direct application in machine learning models for tasks such as link prediction, node classification or community detection, and obviates the need for cumbersome task-specific feature engineering from the input networks.
They have been successfully applied in many domains
such as recommender systems~\cite{Shi2019HERec,Ying2018GraphCN,Hu2018LeveragingMB}, natural language processing~\cite{tu2017cane,Hu2019HeterogeneousGA,tang2015pte} and computational biology~\cite{Su2020NetworkEI,Zitnik2018,Nelson2019ToEO}. 

Many network embedding methods have been proposed for homogeneous networks where nodes and edges are both of single type; well-known examples include Node2vec~\cite{Grover2016node2vec}, DeepWalk~\cite{Perozzi2014DeepWalk}, SDNE~\cite{Wang2016StructuralDN} and LINE~\cite{Tang2015LINE}.
Many real-world interactions are multimodal and multi-typed that give rise to heterogeneous networks where nodes and/or edges can be of different types.
Representation learning methods for such networks have also been widely studied, e.g.,
Metapath2vec~\cite{Dong2017metapath2vec}, HAN~\cite{Wang2019HAN}, HetGNN~\cite{Zhang2019HGN}. 

\begin{figure}[!t]
\centering
\includegraphics[width=1.0\columnwidth]{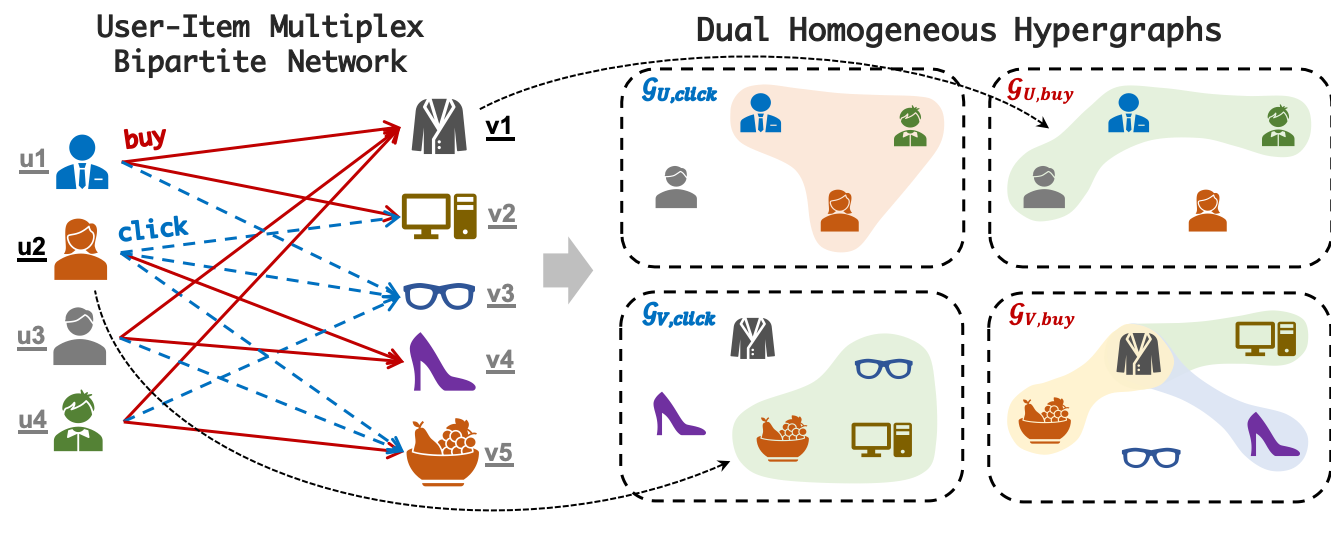}
\caption{(left) User-item multiplex bipartite network and (right) dual homogeneous hypergraphs. E.g., the user $u_2$ in the user-item multiplex bipartite network corresponds to a hyperedge that connects $v_2$, $v_3$ and $v_5$ in the homogeneous hypergraph $\mathcal{G}_{V,click}$ because these three items have been clicked by the same user. Similarly, an item $v_1$ corresponds to a hyperedge that connects $u_1$, $u_3$ and $u_4$ in the homogeneous hypergraph $\mathcal{G}_{U,buy}$ because these three users buy the same item.}
\label{fig1}
\end{figure}

The bipartite network has a specific topology, consisting of two node types (see Figure \ref{fig1}) from different domains, containing inter-domain interactions and no intra-domain interactions.
Essentially representing matrices, such networks are ubiquitous in a variety of contexts.
While general representation learning methods can be applied on such networks, it has been shown that they yield suboptimal representations because many specific characteristics of bipartite networks, such as the two distinct node types and the power-law distribution of node degrees 
may not be modeled well.
As a result, several representation learning methods have been developed specifically for bipartite networks, e.g., BiNE~\cite{BiNE2018}, BGNN~\cite{He2019BipartiteGN}, BiANE~\cite{Huang2020BiANEBA}, FOBE and HOBE~\cite{Sybrandt2019FOBEAH}. 
However, these methods do not model heterogeneous interactions or multiple edges types in bipartite networks. 
Such networks, also called multiplex bipartite networks, model many real-life scenarios.
For example, users and items in an e-commerce platform form a multiplex bipartite network where users have different kinds of interactions (click, inquiry, buy) with items. 

The fundamental challenge in any network representation learning method is to learn the similarities or correlations between nodes, from all the given information about the topology, multiple node and edge types and, if provided, the attributes; and preserve the correlations at the latent level in the embeddings~\cite{Chen2018PME}.
In fact, various network embedding techniques are equivalent to factorization of a node similarity matrix with suitable definitions of similarities~\cite{liu2019general} or tensor factorization~\cite{Yin2017SPTF}.
In bipartite networks, edges provide information about inter-domain node correlations only, while one has to learn intra-domain correlations indirectly.
When node attributes are given, attribute and topology based correlations, representing two different modalities, have to be learnt jointly \cite{Huang2020BiANEBA}.
With the addition of multiple edge types in a multiplex bipartite network, we have more information to model the correlations but 
generalizing the node similarities using heterogeneous edges with potentially distinct distributional and structural properties can be challenging. 

The problem is exacerbated by sparsity of edges and imbalance of distributions of node and edge types in most real-world data.
For instance, consider the Alibaba dataset containing user behavior logs from Alibaba.com (more data details are in Section 5.1).
There are 6,054 users and 16,595 items and the average degree of users and items are 7.55 and 2.76 respectively. 
Each user, on average, interacts with less than 0.1\% items.
Figure~\ref{fig9} shows the clearly distinguishable degree distributions of users and items, with the item distribution having a steeper decline.
Further, each edge type can be present in different proportions and sparsity levels, 
e.g. varying from 25,180 `click' edges 
to 4,429 `contact' edges 
(Figure~\ref{fig10}.a). 


In this paper, we address these challenges by designing a new representation learning model for multiplex bipartite networks.
A key step in our approach is to transform the input into two sets of hypergraphs, a set each for a domain in the bipartite network and a hypergraph for each edge type within a set.
The transformation is scalable since the total number of edges in the hypergraphs is proportional to the number of nodes in the input and to the number of edge types.
A hypergraph generalizes the notion of an edge in simple graphs to a hyperedge which can connect more than two nodes.
This transformation effectively serves many purposes.
It naturally models sparse and heterogeneous interactions in the input, e.g., in the e-commerce network, multiple items naturally form a hyperedge with a user if they are bought (clicked, or inquired) by the same user, and similarly, multiple users can be connected by a hyperedge to an item (see Figure \ref{fig1}).
The homogeneity in these hypergraphs allows us to leverage hypergraph convolutional operators to learn rich representations, capturing local and higher-order structural relationships.
However, this alone is not sufficient to capture inter- and intra-domain correlations in the bipartite network.
To model these correlations and tackle the imbalance problem in both edge and node types, we design additional 
intra- and inter-message passing strategies that enable information  
exchange within and across domains.
Further, our method can also incorporate information from node attributes when provided as inputs.

Our model, called \textbf{Dual} \textbf{H}yper\textbf{G}raph \textbf{C}onvolutional \textbf{N}etworks (\textbf{DualHGCN}), is evaluated through extensive experiments.
On node classification and link prediction tasks, DualHGCN significantly outperforms fourteen state-of-the-art network embedding methods on four real datasets. 
Our experiments also demonstrate the superiority of our model with respect to robustness to varying sparsity levels, node attribute initialization strategies and handling of imbalanced classes.

\section{Related works}
\noindent\textbf{Homogeneous Network Embedding.} Homogeneous networks 
contain a single type of nodes and edges, and thus the sum of node types and edge types is equal to 2. 
Many 
approaches have been proposed 
for homogeneous network embedding methods such as DeepWalk~\cite{Perozzi2014DeepWalk}, Node2vec~\cite{Grover2016node2vec}, LINE~\cite{Tang2015LINE}, SDNE~\cite{Wang2016StructuralDN}, GCN~\cite{kipf2017gcn}, GraphSAGE~\cite{Hamilton2017graphsage} and GAT~\cite{velickovic2018gat}. 
However, these methods 
do not explicitly model bipartite structure and multiple edge types.

\noindent\textbf{Heterogeneous Network Embedding.} A network is called heterogeneous if the sum of node types and edge types is larger than 2. Although multiplex bipartite networks can be viewed as special cases of heterogeneous networks, existing heterogeneous network embedding methods (e.g., Metapath2vec~\cite{Dong2017metapath2vec}, HAN~\cite{Wang2019HAN}, HetGNN~\cite{Zhang2019HGN}, and DyHATR~\cite{Xue2020ModelingDH}) are not tailored to make use of the bipartite topology information and may result in sub-optimal embedding for multiplex bipartite networks. 
For example, Metapath2vec~\cite{Dong2017metapath2vec} uses the meta-path-guided random walk strategy but ignores the difference between explicit and implicit relations and thus becomes suboptimal for network embedding for bipartite networks~\cite{BiNE2018}. Similarly, the node-level and edge-level attention models in DyHATR~\cite{Xue2020ModelingDH} neglect the unique characteristics of the bipartite network, and also do not work well with increasing sparsity.

\noindent\textbf{Bipartite Network Embedding.} Different from multiplex bipartite networks, simple bipartite networks contain two types of nodes and a single type of edges. 
Several bipartite network embedding methods have been proposed, including BiNE~\cite{BiNE2018,BiNE2019}, BGNN~\cite{He2019BipartiteGN}, BiANE~\cite{Huang2020BiANEBA}, FOBE and HOBE~\cite{Sybrandt2019FOBEAH}. 
BiNE first performs biased random walks to generate node sequences and then uses a joint optimization strategy to preserve both explicit and implicit information within bipartite networks simultaneously. As a random walk-based approach, the performance of BiNE deteriorates when the bipartite network becomes sparse. Moreover, BiNE neglects the inherent difference between two types of nodes and models all nodes in the same way. 
BGNN respects the distinction between two types of nodes and proposes a cascaded and unsupervised learning method, which contains inter-domain message passing and intra-domain distribution alignment, to model both same-domain information and cross-domain correlations simultaneously. 
FOBE and HOBE also distinguish two types of nodes and fit embeddings by optimizing nodes of each type separately. They adopt two sampling strategies to generate indirect node-pair sets, including sampling direct and observed pairs (FOBE) and sampling higher-order pairs using algebraic distance (HOBE). 
BiANE is an attributed bipartite network embedding method that differs from the previous three models. It models structural information of the bipartite network through intra- and inter-partition proximity, and integrates attributes and topological structure of networks by a latent correlation model. 

All these existing methods have been designed for bipartite networks where all edges are of the same type and their performance on multiplex bipartite networks suffer without explicit modeling of heterogeneous edge types (as seen in our experiments). Besides, BiNE, FOBE and HOBE cannot capture the inherent attributes of nodes in the bipartite network.

\noindent\textbf{Hypergraph Embedding.} Hypergraph embedding is gaining popularity because of its effectiveness in modeling complex structures within networks. 
A hypergraph is a generalization of a simple graph in which an hyperedge can connect more than two nodes. 
HyperGCN~\cite{Yadati2019HyperGCNAN} decomposes each hyperedge into a collection of node pairs and translates the hypergraph learning tasks into the embedding problem on simple graphs. 
Several homogeneous hypergraph embedding methods have been proposed. 
HGNN~\cite{Feng2019HypergraphNN}, HyperRec~\cite{Wang2020NextitemRW} and HCHA~\cite{Bai2019HypergraphCA} introduce a spectral convolution operator into the hypergraph learning model and capture higher-order structures in hypergraphs.
%
MGCN~\cite{Chen2020MultilevelGC} considers both local and hypergraph level graph convolutions and is able to capture wider and richer network information for network embedding.
Different from previous approaches, HNHN~\cite{Dong2020HNHNHN} introduces a flexible normalization scheme and a hypergraph convolutional model with nonlinear activation neurons on both hypernodes and hyperedges. 
%
Three recent approaches have been proposed to model heterogeneous hypergraphs, e.g., DHNE~\cite{Tu2018StructuralDE}, Hyper-SAGNN~\cite{Zhang2020HyperSAGNN}, and HWNN~\cite{Sun2020HWNN}. 
However, most previous hypergraph embedding methods have been designed for supervised tasks, and cannot be directly used for obtaining network embeddings.
Besides, hypergraph convolution networks do not specifically model multiplex edges and imbalanced degrees.

\section{Preliminaries}
In this section, we define a multiplex bipartite network and its vertex embedding.
Table~\ref{notations} provides a summary of frequently used symbols in the paper.

\begin{table}[!t]
\caption{Summary of notations and descriptions.}\label{notations}
\centering
\begin{tabular}{c|c}
\toprule
Notations & Descriptions\\
\midrule
$G=(U,V,E,X)$ & Multiplex Bipartite Network (MBN) \\
$U$, $V$ & Two types of node sets in MBN \\
$E = E^{1} \cup \ldots \cup E^{k}$ & A set of edges in MBN \\
$X=\{X_u,X_v\}$ & Two feature sets of $U$ and $V$ in MBN \\
$k$ & The number of edge types \\
$\mathbf{G}_U, \mathbf{G}_V$ & Two homogeneous hypergraph sets \\
$\mathcal{G}_{U,i}$, $\mathcal{G}_{V,j}$ & A homo-hypergraph from $U$, $V$ \\
$\mathbf{H}$ & The incidence matrix of hypergraph \\
$\mathbf{D}$ & The diagonal matrices of node degree \\
$\mathbf{B}$ & The diagonal matrices of hyperedge degree \\
$\mathbf{W}$ & The diagonal identity matrix \\
$\mathbf{X}_{U,i}$, $\mathbf{X}_{V,j}$ & The learned features of $\mathcal{G}_{U,i}$, $\mathcal{G}_{V,j}$ \\
$\mathbf{P}$, $\mathbf{Q}$ & Learnable weight matrix\\
$\sigma(\cdot)$ & Activation function \\
$\mathbf{Z}=\{\overline{\mathbf{X}}_{U},\overline{\mathbf{X}}_{V}\}$ & Final embeddings of $\mathbf{G}_U, \mathbf{G}_V$ \\
$n$ & Negative sample parameter \\
$d$ & Dimension of the embedding \\
\bottomrule
\end{tabular}
\end{table}

\noindent\textbf{Definition 2.1. Multiplex Bipartite Network.} A Multiplex Bipartite Network $G=(U,V,E,X)$ consists of node sets $U$ and $V$ of different types, and edge sets $E = E^{1} \cup E^{2} \ldots \cup E^{k}$ where $E^{i}$ denotes the $i$-th type of edge. $X=\{X_u,X_v\}$ denotes the features of node sets $U$ and $V$. 

For example, logs of user behavior in Alibaba.com can be represented as a multiplex bipartite network containing two types of nodes (users and items) and several types of edges (e.g., click, enquiry, contact). For each user and item, the logs also contain unique attributes with different dimensions. For instance, attributes of users contain country, gender, search logs, etc. In contrast, Items usually have attributes, such as category, price, search counts, visit counts, buy logs, etc.

\noindent\textbf{Definition 2.2. Multiplex Bipartite Network Embedding.} Given a multiplex bipartite network $G=(U,V,E,X)$, its embedding is a $d$-dimensional feature $\overline{\mathbf{X}}_{U}\in\mathbb{R}^{|U|\times d}$, $\overline{\mathbf{X}}_{V}\in\mathbb{R}^{|V|\times d}$ for each node in $U$ and $V$, where $d\ll|U|$ and $d\ll|V|$, that captures information of both the global topological structure and node attributes. 

We define the sparsity of a multiplex bipartite network as $\langle\mathcal{S}\rangle=1-\frac{|E|}{|U|\times |V|}$. 
Many real-world multiplex bipartite networks are extremely sparse, i.e., $|E|\ll|U|\times|V|$. 
For instance, in the  Alibaba dataset $\langle\mathcal{S}\rangle = 99.95\%$. 

\begin{figure*}[htbp]
\centering
\includegraphics[width=2.0\columnwidth]{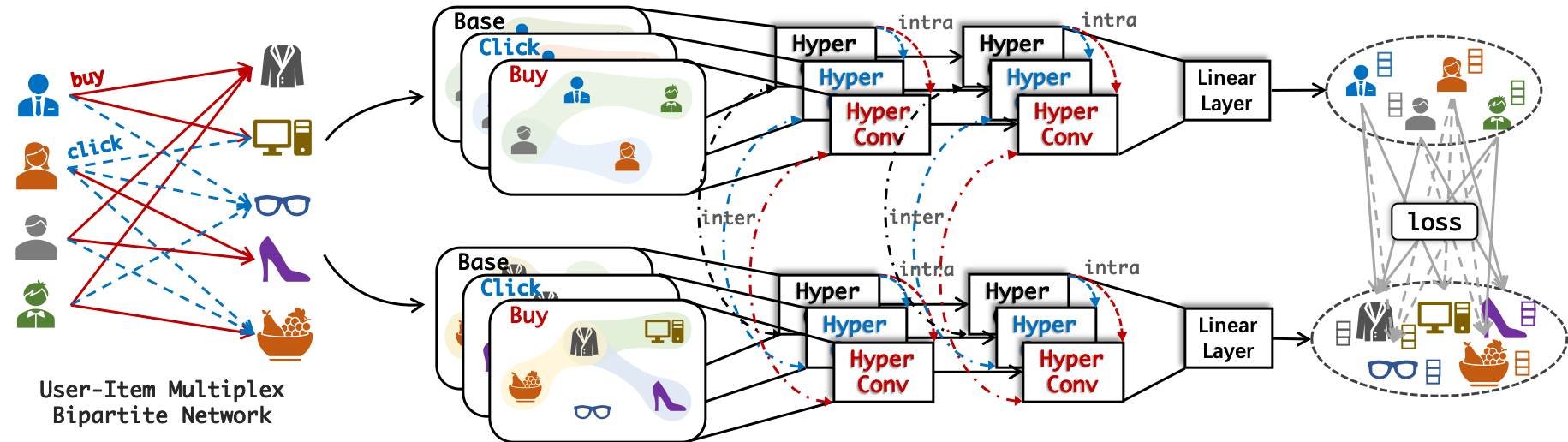}
\caption{The Overall framework of the proposed DualHGCN method.}
\label{framework}
\end{figure*}

\section{Methodology}
Given an input multiplex bipartite network, we first transform it into two sets of homogeneous hypergraphs.
Our model architecture comprises a hypergraph convolutional network that assumes these dual homogenous hypergraphs as inputs, with additional inter- and intra-message passing layers to enable information sharing across the networks.
Finally, the entire model is trained using a gradient descent based optimizer.
The next four subsections provide more details.
Figure~\ref{framework} shows an overview of the entire method.

\subsection{Dual Homo-Hypergraphs Construction}

We now show how to transform a multiplex bipartite network into two sets of homogeneous hypergraphs (dual homo-hypergraphs). We construct two sets of homogeneous hypergraphs $\mathbf{G}_U, \mathbf{G}_V$, from node sets $U,V$, respectively, as follows:
\begin{equation}
    \mathbf{G}_U =\{\mathcal{G}_{U,base},\mathcal{G}_{U,1},...,\mathcal{G}_{U,k}\}, 
    \mathbf{G}_V = \{\mathcal{G}_{V,base},\mathcal{G}_{V,1},...,\mathcal{G}_{V,k})\},
\end{equation}
where $\mathcal{G}_{U,i}=\{U,\mathcal{E}_{U,i}\}$, $\mathcal{G}_{V,j}=\{V,\mathcal{E}_{V,j}\}$, and $\mathcal{E}_{U,i}$ and $\mathcal{E}_{V,j}$ denote hyperedges in $\mathcal{G}_{U,i}$ and $\mathcal{G}_{V,j}$ respectively. 
Note that all the homogeneous hypergraphs in $\mathbf{G}_U$ share the same node set $U$ while all the homogeneous hypergraphs in $\mathbf{G}_V$ share the same node set $V$. For a node $v \in V$, a hyperedge is introduced in $\mathcal{E}_{U,i}$ of $\mathcal{G}_{U,i}$ which connects to $\{u| u \in U, (u,v) \in E^i\}$, i.e., the vertices in $U$ that are directly connected to $v$ by $E^i$. Similarly, for a node $u \in U$, a hyperedge is introduced in $\mathcal{E}_{V,j}$ of $\mathcal{G}_{V,j}$ which connects to $\{v| v \in V, (u,v) \in E^j\}$, i.e., the vertices in $V$ that are directly connected to $u$ by $E^j$. 

Refer to Figure~\ref{fig1} for an example. In the user-item multiplex bipartite network, the user $u_2$ clicks three items ($v_2$, $v_3$ and $v_5$), which corresponds to a hyperedge that connects these three items in the homogeneous hypergraph $\mathcal{G}_{V,click}$. Similarly, the item $v_1$ is bought by three users ($u_1$, $u_3$ and $u_4$) which corresponds to a hyperedge that connects these three users in the homogeneous hypergraph $\mathcal{G}_{U,buy}$. 

Two special homogeneous hypergraphs $\mathcal{G}_{U,base} \in \mathbf{G}_U$ and $\mathcal{G}_{V,base} \in \mathbf{G}_V$ are defined as $\mathcal{G}(U,\bigcup\limits_{i=1}^{k} \mathcal{E}_{U,i}) $ and $\mathcal{G}(V,\bigcup\limits_{j=1}^{k} \mathcal{E}_{V,j})$, respectively. Note that the cardinalities of hyperedge sets in the constructed hypergraphs are: $|\mathcal{E}_{U,i}| \leq |V|$, $|\mathcal{E}_{V,j}| \leq |U|$, $|\mathcal{E}_{U,base}| \leq k|V|$ and $|\mathcal{E}_{V,base}| \leq k|U|$ for $1 \leq i, j \leq k$.
The total number of hyperedges in the dual homo-hypergraphs is proportional to the number of nodes and edge types in the input network: $O(k(|U| + |V|))$. Thus, the transformation easily scales to large inputs.



\subsection{Hypergraph Convolutional Networks}
Note that the hypergraphs that we constructed from a multiplex bipartite network are homogeneous and now we can apply hypergraph convolutions on them to learn  representations. 
Graph convolutional network~\cite{kipf2017gcn} has been widely used in modeling simple networks. Recent hypergraph convolutional operators have borrowed ideas from the spectral theory on simple graphs and achieved good performance in hypergraph embedding (e.g., HGNN, HCHA, HyperGCN and MGCN). We briefly describe two classical hypergraph convolutional operators used in our model. 

Simple graphs use the adjacency matrix $A$ to represent edges, whereas hypergraphs introduce the incidence matrix $H$ to describe the relationship between nodes and hyperedges. 
Given a 
homo-hypergraph $\mathcal{G}_{U,i}=(U,\mathcal{E}_{U,i})$ where $i\in\{base,1,...,k\}$, $k$ is the number of edge types, 
the incidence matrix 
of $\mathcal{G}_{U,i}$ is defined as:
\begin{equation}
    \mathbf{H}_{U,i}(u,e)=\begin{cases}
    1,  & \mbox{if }u \mbox{ is incident to } e, e \in \mathcal{E}_{U,i} \\
    0,  & \mbox{otherwise,}
    \end{cases}
\end{equation}
where $\mathcal{E}_{U,i}$ denotes the set of hyperedges in $\mathcal{G}_{U,i}$, $\mathbf{H}_{U,i}\in{\mathbb{R}^{|U|\times|\mathcal{E}_{U,i}|}}$, and $i\in\{base,1,...,k\}$ denotes the constructed homo-hypergraph $i$. 
Similarly we define the incidence matrix $\mathbf{H}_{V,j}$ for $\mathcal{G}_{V,j}$. 
Let $\mathbf{D}_{U,i}\in\mathbb{R}^{|U|\times|U|}$ and $\mathbf{B}_{U,i}\in\mathbb{R}^{|\mathcal{E}_{U,i}|\times|\mathcal{E}_{U,i}|}$ denote diagonal matrices of the node degree and the hyperedge degree respectively, where $\mathbf{D}_{U,i}(u,u)=\sum_{e\in\mathcal{E}_{U,i}}\mathbf{H}_{U,i}(u,e)$ and $\mathbf{B}_{U,i}(e,e)=\sum_{u\in U}\mathbf{H}_{U,i}(u,e)$. 

Two hypergraph spectral convolutional operators, symmetric hypergraph convolution (sym) and the asymmetric hypergraph convolution (asym), are used to learn embeddings of each hypergraph in our model. 
For a simple graph, the convolutional operator can be formulated as $X^{l+1}=\sigma(A\cdot X^{l}P^{l})$, where $X$ is the feature matrix, $A$ is the adjacency matrix and $P$ is the learnable weight matrix. Because the incidence matrix $\mathbf{H}$ denotes the relationship between nodes and hyperedges, we use $\mathbf{H}\mathbf{W}\mathbf{H}^{\top}$ to measure the pairwise relationships between nodes in the same homogeneous hypergraph, where $\mathbf{W}$ is the weight matrix that assigns weights for all hyperedges. Usually, we initialize the weight matrix $\mathbf{W}$ with the identity matrix yielding equal weights for all hyperedges. Thus, the intuitive hypergraph convolutional operator can be formulated as:
\begin{equation}
    \mathbf{X}^{l+1}=\sigma(\mathbf{H}\mathbf{W}\mathbf{H}^{\top}\cdot\mathbf{X}^{l}\mathbf{P}^{l})
\end{equation}
However, the previous hypergraph convolutional operator may change the scale of the feature vectors $\mathbf{X}$ by adding layers of convolutional operators (multiplication with $\mathbf{H}\mathbf{W}\mathbf{H}^{\top}$). To constrain the number of parameters and decrease the number of matrix multiplications, and thereby avoid the overfitting problem, GCN introduces a renormalization trick, $X^{l+1}=\sigma((I+D^{-1/2}AD^{-1/2})\cdot X^{l}P^{l})=\sigma(\widetilde{D}^{-1/2}\widetilde{A}\widetilde{D}^{-1/2}\cdot X^{l}P^{l})$, where $\widetilde{A}=A+I$, $\widetilde{D}_{ii}=\sum_{j}\widetilde{A}_{ij}$, $I$ is the identity matrix 
and $D$ is the node degree matrix of a simple graph. Similarly, the symmetric normalization version of hypergraph convolutional operator for $\mathcal{G}_{U,i}$ can be defined as:

\begin{equation}\label{eq:hyperconv-sym}
\mathbf{X}_{U,i}^{l+1}=\sigma(\mathbf{D}_{U,i}^{-1/2}\mathbf{H}_{U,i}\mathbf{W}_{U}\mathbf{B}_{U,i}^{-1}\mathbf{H}_{U,i}^{\top}\mathbf{D}_{U,i}^{-1/2}\cdot\mathbf{X}_{U,i}^{l}\mathbf{P}_{U,i}^{l}),
\end{equation}
and, the asymmetric hypergraph convolutional operator for $\mathcal{G}_{U,i}$ can be defined as:
\begin{equation}\label{eq:hyperconv-asym}
\mathbf{X}_{U,i}^{l+1}=\sigma(\mathbf{D}_{U,i}^{-1}\mathbf{H}_{U,i}\mathbf{W}_{U}\mathbf{B}_{U,i}^{-1}\mathbf{H}_{U,i}^{\top}\cdot\mathbf{X}_{U,i}^{l}\mathbf{P}_{U,i}^{l}),
\end{equation}

where $\sigma$ denotes the nonlinear activation function (i.e., ReLU function in our model), $\mathbf{X}_{U,i}^{l}\in\mathbb{R}^{|U|\times d_l}$ is the feature of the $l$-th layer, $\mathbf{W}_U\in\mathbb{R}^{|V|\times|V|}$ is the identity matrix, and $\mathbf{P}_{U,i}^{l}\in\mathbb{R}^{d_{l}\times d_{l+1}}$ denotes the learnable filter matrix, $d_{l}$ and $d_{l+1}$ are the dimensions of the $l$-th and $(l+1)$-th layers respectively. 

Similar hypergraph convolutional operators are also applied to learn features from $\mathcal{G}_{V,i}$. 
Therefore, for dual homo-hypergraphs $\mathcal{G}$, we can learn features from each homo-hypergraph ($\mathcal{G}_{U,i}$ or $\mathcal{G}_{V,j}$) independently through the above hypergraph convolutional operators (Eqs. \ref{eq:hyperconv-sym} and \ref{eq:hyperconv-asym}). Thus, we obtain the low-dimensional node representations $\{\mathbf{X}_{U,base}, \mathbf{X}_{U,1},...,\mathbf{X}_{U,k}\}$ and $\{\mathbf{X}_{V,base},\mathbf{X}_{V,1},...,\mathbf{X}_{V,k}\}$. 

These hypergraph convolutional operators can model each homo-hypergraph, but cannot handle the problem of multiplex edges and topological imbalance. Thus, as described in the following section, we add new layers to enable intra- and inter- message-passing.

\subsection{Message-passing Strategies}
In the previous section, the multiplex bipartite network is transformed into independent hypergraphs ($\mathcal{G}_{U,i}$ or $\mathcal{G}_{V,j}$) that correspond to each edge type. There may be information loss with respect to each node in the hypergraphs because correlations between different edge types have not been modeled. 
Take the e-commerce platform for an example, a user is more likely to `buy' an item after this user `inquiries' this item or similar items, but the current embeddings consider `buy' and `inquiry' independently as they are two different edge types between user and item. Therefore, we introduce the following intra-message passing strategy to promote information sharing among $\{\mathbf{X}_{U,base}, \mathbf{X}_{U,1},...,\mathbf{X}_{U,k}\}$ and among $\{\mathbf{X}_{V,base}, \mathbf{X}_{V,1},...,\mathbf{X}_{V,k}\}$.

\textbf{Intra-message passing.} 
As $\mathcal{G}_{U,base}$ contains information aggregated from all $\mathcal{G}_{U,i}$, we incorporate information from the learned $\mathbf{X}_{U,base}$ into each $\mathbf{X}_{U,i}$. The iterative formula of the intra-message passing strategy (from $l$-th layer to $(l+1)$-st layer) is defined as:
\begin{equation}
    \mathbf{X}_{U,i}^{l+1}=\sigma(\Theta_{U,i}\cdot\mathbf{X}_{U,i}^{l}\mathbf{P}_{U,i}^{l}+\mathbf{X}_{U,base}^{l}\mathbf{Q}_{U,base}^{l})
\end{equation}
where $\Theta_{U,i}=\mathbf{D}_{U,i}^{-1/2}\mathbf{H}_{U,i}\mathbf{W}_{U}\mathbf{B}_{U,i}^{-1}\mathbf{H}_{U,i}^{\top}\mathbf{D}_{U,i}^{-1/2}$ for symmetric convolutional operators or $\Theta_{U,i}=\mathbf{D}_{U,i}^{-1}\mathbf{H}_{U,i}\mathbf{W}_{U}\mathbf{B}_{U,i}^{-1}\mathbf{H}_{U,i}^{\top}$ for asymmetric convolutional operators (from  Eqns \ref{eq:hyperconv-sym} and \ref{eq:hyperconv-asym}) 
and $\mathbf{Q}_{U,base}^{l}$ denotes the learnable transform matrix. 

As discussed in Section 4.1, when a multiplex bipartite network is transformed into homogeneous hypergraphs, a node $u$ in $U$ corresponds to a hyperedge $e \in \mathcal{E}_{V,j}$ in the hypergraph $\mathcal{G}_{V,j}$. However, the hypergraph convolutional operators in the previous section neglects the above correspondence between nodes and hyperedges and thus may result in suboptimal embeddings. 
Therefore, we introduce the following inter-message passing to reinforce similar properties between $\mathbf{X}_{U,i}$ and $\mathbf{X}_{V,i}$ with respect to the same $i$-th edge type. 

\textbf{Inter-message passing.} 
We propose an inter-message passing strategy which fuses the features $\mathbf{X}_{U,i}$ and $\mathbf{X}_{V,j}$ (from $l$-th layer to $l+1$-th layer) and is given by:
\begin{equation}
    \mathbf{X}_{U,i}^{l+1}=\sigma(\Theta_{U,i}\cdot\mathbf{X}_{U,i}^{l}\mathbf{P}_{U,i}^{l}+\mathbf{H}_{V,i}^{\top}\mathbf{X}_{V,i}^{l}\mathbf{Q}_{V,i}^{l})
\end{equation}
\begin{equation}
    \mathbf{X}_{V,j}^{l+1}=\sigma(\Theta_{V,j}\cdot\mathbf{X}_{V,j}^{l}\mathbf{P}_{V,j}^{l}+\mathbf{H}_{U,j}^{\top}\mathbf{X}_{U,j}^{l}\mathbf{Q}_{U,j}^{l})
\end{equation}
where $\mathbf{H}_{U,i}$ and $\mathbf{H}_{V,j}$ denote the incidence matrix, $\mathbf{Q}_{U,i}^{l}$ and $\mathbf{Q}_{V,j}^{l}$ denote the learnable transform matrix. After $t$ iterations, the embeddings of nodes $U$ and $V$ can be formulated as $\mathbf{X}_{U}^t=\{\mathbf{X}_{U,base}^t,\mathbf{X}_{U,1}^t,$ $...,\mathbf{X}_{U,k}^t\}$, $\mathbf{X}_{U,i}^t\in\mathbb{R}^{|U|\times d_t}$, and $\mathbf{X}_{V}^t=\{\mathbf{X}_{V,base}^t,\mathbf{X}_{V,1}^t,...,\mathbf{X}_{V,k}^t\}$, $\mathbf{X}_{V}^t\in\mathbb{R}^{|V|\times d_t}$, where $d_t$ is the dimension of final embeddings and $k$ is the number of edge types. 
Then, we concatenate these learned features after $t$ layers training for two types of nodes, $\mathbf{X}_{U}^t$ and $\mathbf{X}_{V}^t$, and 
pass them to a linear layer to obtain the final embeddings:
\begin{equation}
    \overline{\mathbf{X}}_U=\mathbf{X}_U^t\cdot\mathbf{W}_U+b_U, \quad 
    \overline{\mathbf{X}}_V=\mathbf{X}_V^t\cdot\mathbf{W}_V+b_V,
\end{equation}
where $\mathbf{W}_U, \mathbf{W}_V\in \mathbb{R}^{((k+1)*d_t)\times d_t}$, and $b_U, b_V\in\mathbb{R}^{d_t}$ are trainable parameters, and $\overline{\mathbf{X}}_{U}\in\mathbb{R}^{|U|\times d_t}$, $\overline{\mathbf{X}}_{V}\in\mathbb{R}^{|V|\times d_t}$. Thus, after training the network, we can obtain final embeddings: $\mathbf{Z}=\{\overline{\mathbf{X}}_{U},\overline{\mathbf{X}}_{V}\}$.

\subsection{Optimization}
To learn the weights of DualHGCN, we maximize the probability of positive edges (existing edges in the multiplex bipartite network) and minimize the probability of negative ones (unseen edges): 
\begin{equation}\label{eq:lossfunc}
\begin{split}
 & L=\sum\limits_{(u,v)\in E} \bigg[\lambda\cdot\log\sigma(\mathbf{Z}_{u}^{\top}\mathbf{Z}_{v})+
  (1-\lambda)\cdot\sum\limits_{i=1}^{n} \Big(\mathbb{E}_{{u_i}\sim P(u)} \\
 & \log(1-\sigma(\mathbf{Z}_{u}^{\top}\mathbf{Z}_{u_i}))+
  \mathbb{E}_{{v_i}\sim P(v)}\log(1-\sigma(\mathbf{Z}_{v}^{\top}\mathbf{Z}_{v_i}))
  \Big)
  \bigg]
\end{split}
\end{equation}
where $\sigma$ is the sigmoid activation function, $\lambda$ denotes the weight to balance the importance between positive and negative samples, $P(u)$ denotes the negative candidate nodes distribution of $u$, and $n$ is the number of the negative samples. The existing edges in the multiplex bipartite network are treated as positive samples. For each positive pairwise edge $(u,v)$, we randomly sample $n$ negative edges incident to node $u$ and $v$, respectively. The pseudocode for DualHGCN is shown in Algorithm~\ref{alg:dualhgcn}.

\begin{algorithm}
\caption{The DualHGCN algorithm.}
\label{alg:dualhgcn}
\KwIn{Multiplex bipartite network $G=(U,V,E,X)$, number of iterations $t$, number of negative samples per positive sample $n$, the initial features $X$, initial parameters\;}
\KwOut{Node Embedding $\mathbf{Z}$}
\textbf{Model Construction:}

Construct dual homo-hypergraphs $\mathcal{G}$\; 
Add spectral convolutional layers on Base homo-hypergraph (Eq. (4) or (5)), $\mathbf{X}_{U,base}^i$, $\mathbf{X}_{V,base}^i$\;
\For{each edge-type $j\in [1,2,...,k]$}{
Add intra and inter-message passing layers (Eq. (6), (7) and (8)), $\mathbf{X}_{U,j}^i$, $\mathbf{X}_{V,j}^i$\;
}
Add linear layer to outputs of hypergraph convolutions\;
\textbf{Optimization:}

Initialize Embeddings $\mathbf{Z}$ with initial features\;
Randomly sample $n$ negative edges for each positive edge\;
Optimize loss (Eq. (\ref{eq:lossfunc})) via gradient descent ($t$ iterations)\;
\textbf{return} $\mathbf{Z}$;
\end{algorithm}

\section{Experiments}
We benchmark our proposed model with several baselines to validate the effectiveness of DualHGCN for unsupervised multiplex bipartite network representation learning. Specifically, we investigate the following questions in these carefully designed experiments:
\begin{itemize}
    \item[\textbf{Q1}] How does DualHGCN perform in predicting unknown interactions or user behaviors (i.e., the link prediction task)?
    \item[\textbf{Q2}] How does DualHGCN perform in classifying items according to user behaviors (i.e., the node classification task)?
    \item[\textbf{Q3}] How do the inter- and intra-message passing strategy contribute to final unsupervised embeddings of DualHGCN?
    \item[\textbf{Q4}] How do the multiple types of edges and the sparsity of networks affect the performance of DualHGCN?
    \item[\textbf{Q5}] How sensitive is the performance of DualHGCN to its parameter settings?
\end{itemize}

\subsection{Datasets} 
We use four real-world datasets in our experiments. 
Their detailed statistics are given in Table~\ref{tab2}.

\noindent\textbf{DTI.\footnote{https://drugtargetcommons.fimm.fi}} This Drug-Target Interactions bipartite network was randomly sampled from the data in the Drug Target Commons platform~\cite{Tang2018DrugTC}. 
The sampled dataset mainly contains two types of nodes (drugs and targets) and five bio-activities (Potency, IC50, KI, Inhibition, and Activity) that form the edges.

\noindent\textbf{Amazon.\footnote{http://jmcauley.ucsd.edu/data/amazon}} This 
dataset is a heterogeneous non-bipartite network ~\cite{Cen2019RepresentationLF}. We follow the strategy in BGNN~\cite{He2019BipartiteGN} to process this dataset to derive a multiplex bipartite network. This multiplex bipartite network contains two types of edges and two types of nodes, where the attributes of nodes include the price, sales-rank, brand, and category.

\noindent\textbf{Alibaba-s} and \textbf{Alibaba.\footnote{https://www.alibaba.com}} These real-world datasets consist of behavior logs of users and items collected from the e-commerce platform Alibaba.com from $1^{\rm st}$ April 2020 to $30^{\rm th}$ April 2020. 
It contains two types of nodes (users and items) and three types of activities (click, enquiry and contact). 
The items are classified into five categories (women's clothing, men's clothing, etc.). 
Alibaba-s is a smaller unattributed dataset, and Alibaba is a multiplex bipartite network where users have attributes such as country, gender and search logs, and items have attributes including the category, price, search counts, visit counts, buy logs, etc. 
We have anonymized all sensitive data, e.g., user and item id, in both Alibaba-s and Alibaba datasets.


\begin{table}[!t]
\centering
\caption{Statistics of four real-world datasets. The sparsity of the multiplex bipartite network: $\langle\mathcal{S}\rangle=1-\frac{|E|}{|U|\times|V|}$.}
\label{tab2}
\begin{tabular}{c|c|c|c|c|c}
\toprule
\multicolumn{2}{c|}{Datasets} & DTI & Amazon & Alibaba-s & Alibaba \\
\midrule
\multirow{2}{*}{\#Nodes} & U & 3,270 & 3,781 & 1,869 & 6,054 \\
\cline{2-6}
 & V & 1,567 & 5,749 & 13,349 & 16,595 \\
\hline
\multicolumn{2}{c|}{\#Edges} & 16,458 & 60,658 & 27,036 & 45,734 \\
\hline
\multicolumn{2}{c|}{\#Edge Types} & 5 & 2 & 3 & 3 \\
\hline
\multirow{2}{*}{\#Features} & U & \multirow{2}{*}{N/A} & \multirow{2}{*}{4} & \multirow{2}{*}{N/A} & 7 \\
\cline{2-2}\cline{6-6}
 & V & & & & 11 \\
\hline
\#Classes & V & N/A & N/A  & 5 & 5 \\
\hline
\multicolumn{2}{c|}{\#$\langle\mathcal{S}\rangle$} & 99.68\% & 99.72\% & 99.89\% & 99.95\% \\
\bottomrule
\end{tabular}
\end{table}

\subsection{Baselines}
We compare DualHGCN with fourteen state-of-the-art algorithms in four categories as listed below.

\textbf{1) Simple Homogeneous Network Embedding.} 
These homogeneous network embedding methods 
ignore the both node-type and edge-type information in the input multiplex bipartite network. They also do not use hypergraphs to generate low-dimensional representations for each node.

\begin{itemize}
    \item Node2vec~\cite{Grover2016node2vec} uses a biased random walk procedure and extends the skip-gram model. 
    \item GraphSAGE~\cite{Hamilton2017graphsage} is an inductive network embedding method which contains several message aggregation strategies to generate features for previously unobserved nodes.
    \item GCN~\cite{kipf2017gcn} proposes a spectral graph convolutional operator to learn both local network structure and features of nodes.
    \item GAT~\cite{velickovic2018gat} uses masked self-attention mechanism to assign different neighbors with different specified weights.
\end{itemize}

\textbf{2) Hypergraph Embedding.} 
For these methods, we use the same strategy mentioned in Section 4.1 to build dual `base' homo-hypergraphs.
They also ignore edge-type information in the inputs.
\begin{itemize}
    \item HGNN~\cite{Feng2019HypergraphNN}: generalizes the spectral convolutional networks 
    to capture high-order structural information. 
    \item HyperGCN~\cite{Yadati2019HyperGCNAN}: decomposes hyperedges of its hypergraphs into a set of node pairs and then uses a simple graph convolutional network to learn decomposed node pairs.
    \item HCHA~\cite{Bai2019HypergraphCA}: uses a spectral convolutional and attention-based method to model multi-hop relationships. 
    \item MGCN~\cite{Chen2020MultilevelGC}: 
    generalizes from simple graph convolutional networks without using spectral convolutions.
\end{itemize}

\textbf{3) Heterogeneous Network Embedding.} 
These methods can model multiple node and edge types but do not explicitly model the bipartite structure of the input multiplex bipartite network.
\begin{itemize}
    \item Metapath2vec++~\cite{Dong2017metapath2vec}: generates meta-path-based random walks on which a heterogeneous skip-gram model is trained.
    \item HAT~\cite{Xue2020ModelingDH}: is a hierarchical attention based heterogeneous network embedding method which uses node-level and edge-level attention to model multiple edges types.
\end{itemize}

\textbf{4) Bipartite Network Embedding.} 
These methods are applied on the `base' bipartite networks constructed in Section 4.1 to derive the final node embeddings. 
\begin{itemize}
    \item BiNE~\cite{BiNE2018}: generates biased random walks and then optimizes to preserve both the explicit and implicit relationships within the bipartite network. 
    \item BGNN~\cite{He2019BipartiteGN}: a cascaded and unsupervised embedding method with a communication strategy between the 
    domains to distinguish between the two types of nodes and promote information sharing across two domains simultaneously.
    \item BiANE~\cite{Huang2020BiANEBA}: an attributed bipartite network embedding method which can model the intra- and inter-partition proximity simultaneously and uses a latent correlation training approach to jointly learn attribute and structure information.
\end{itemize}

As another baseline just the initial features are used, i.e., without any network embedding methods. 
In datasets where node attributes are available, the attributes are used as initial features and in datasets without node attributes, we use a tied autoencoder~\cite{Baldi2011Autoencoders} (where weights across the encoder and decoder are tied) on the adjacency matrix of the multiplex bipartite network to generate initial features.

\subsection{Experimental Settings}
%

\textbf{Link Prediction.} We randomly sample 50\% of the edges as the training set and the remaining edges are treated as the test set. 
The network embedding methods are run on the subgraph formed from training set edges only.
For each edge in the test set, embeddings of the incident nodes (learnt from the training set) are used as features. 
5-fold cross validation is used on the test set edges to evaluate the Logistic Regression classifier performance. 
The entire procedure is repeated 5 times to obtain different random samples of train and test sets. 
Mean and standard deviation values of the classification evaluation metrics are reported. 
We use the area under the ROC curve (AUROC) and the area under the precision-recall curve (AUPRC) as evaluation metrics and Logistic Regression as the classifier. 

\textbf{Node Classification.} 
All the network embedding methods are run on the entire dataset to obtain the node embeddings. 
We use the micro-F1 and macro-F1 as the evaluation metrics and Stochastic Gradient Descent (SGD) classifier. 
5-fold cross validation on the entire data is used to evaluate classifier performance on node classification. We report the mean and standard deviation values. 

\textbf{Statistical Significance.}
To quantify the significance of the improvement achieved by DualHGCN, when compared with baselines, we compute the one-sided Wilcoxon rank-sum p-value~\cite{Gehan1965AGW} between DualHGCN and the next-best results in each experiment. 

\textbf{Parameter Settings.} 
The dimensions of initial features (for both with and without attributes) and final embeddings are all empirically set to be 32. 
%
We run GraphSAGE with different aggregators (e.g., mean, lstm, and pooling) and show the best results. 
In Metapath2vec, we use `U-V-U' as the meta-path to model the `base' bipartite network. 
For all baseline methods, we optimize their models with different parameters and report the best performance scores.

For our method, DualHGCN, the default number of layers is 2. The number of negative samples is different from distinct datasets and ranges in $\{1, 2, 3\}$. 
The Adam optimizer is used in our model to optimize parameters via backpropagation. 
When the symmetric version of the hypergraph convolutional operator (Eq. \ref{eq:hyperconv-sym}) is used in DualHGCN, we call the method DualHGCN-sym and when the asymmetric operator (Eq. \ref{eq:hyperconv-asym}) is used, we call the method DualHGCN-asym.
The classifiers used for link prediction and node classification, and evaluation metrics are all from the scikit-learn library~\cite{scikit-learn}.
All codes, data and experimental settings of the DualHGCN model are freely available\footnote{https://github.com/xuehansheng/DualHGCN}.

\begin{table*}[htbp]
\caption{The AUROC and AUPRC values of DualHGCN and baselines on the task of link prediction($\%$). The initial features for all datasets are adjacency matrix with tied autoencoder.}\label{linkprediction1}
\centering
\begin{threeparttable}
\begin{tabular}{c|c|c|c|c|c|c|c|c}
\toprule
\multirow{2}{*}{Methods} & \multicolumn{2}{c}{DTI} & \multicolumn{2}{|c}{Amazon} & \multicolumn{2}{|c}{Alibaba-s} & \multicolumn{2}{|c}{Alibaba}\\
\cline{2-9}
 & AUROC & AUPRC & AUROC & AUPRC & AUROC & AUPRC & AUROC & AUPRC \\
\midrule
Initial features & 63.43$\pm$0.74 & 72.34$\pm$0.73 & 70.57$\pm$0.55 & 74.50$\pm$0.63 & 67.08$\pm$0.45 & 68.21$\pm$0.67 & 68.06$\pm$0.25 & 71.38$\pm$0.28 \\
\midrule
Node2vec &50.88$\pm$0.37 &57.45$\pm$0.38 &50.30$\pm$0.44 &55.44$\pm$0.47 &50.43$\pm$0.29 &51.42$\pm$0.50 &50.10$\pm$0.49 &51.52$\pm$0.29 \\
GraphSAGE &79.34$\pm$0.39 &82.36$\pm$0.24 &69.99$\pm$0.18 &69.39$\pm$0.30 &64.91$\pm$0.14 &65.76$\pm$0.21 &66.49$\pm$0.09 &60.36$\pm$0.13\\
GCN &56.95$\pm$0.13 &76.00$\pm$0.29 &64.93$\pm$0.12 &77.45$\pm$0.15 &63.08$\pm$0.10 &79.59$\pm$0.15 &56.87$\pm$0.04 &77.66$\pm$0.09 \\
GAT &76.33$\pm$0.25 &80.64$\pm$0.31 &66.70$\pm$0.13 &70.16$\pm$0.15 &53.28$\pm$0.28 &54.29$\pm$0.66 &55.38$\pm$0.30 &54.49$\pm$0.47 \\
\midrule
HGNN & 77.87$\pm$1.07 & 83.57$\pm$1.02 & 80.14$\pm$0.32 & 82.94$\pm$0.17 & 67.07$\pm$0.12 & 69.34$\pm$0.07 & 69.64$\pm$0.15 & 73.50$\pm$0.07 \\
HCHA & 63.77$\pm$1.39 & 69.83$\pm$1.07 & 62.66$\pm$0.72 & 67.84$\pm$0.72 & 63.61$\pm$0.16 & 65.47$\pm$0.18 & 65.84$\pm$0.09 & 68.81$\pm$0.06 \\
MGCN & 50.14$\pm$0.11 & 62.07$\pm$2.86 & 51.84$\pm$1.34 & 62.35$\pm$0.61 & 66.31$\pm$1.19 & 68.60$\pm$1.50 & 51.23$\pm$0.63 & 52.79$\pm$1.20 \\
HyperGCN & 68.99$\pm$1.70 & 77.34$\pm$1.86 & 68.42$\pm$1.02 & 73.78$\pm$0.60 & 63.72$\pm$0.22 & 63.54$\pm$0.17 & 61.38$\pm$1.12 & 65.21$\pm$0.70 \\
\midrule
Metapath2vec++ &85.99$\pm$0.12 &87.73$\pm$0.43 &60.24$\pm$0.17 &63.58$\pm$0.21 &78.85$\pm$0.22 &69.17$\pm$0.37 &65.98$\pm$0.18 &70.97$\pm$0.20 \\
HAT &86.22$\pm$0.14 &87.21$\pm$0.12 &67.26$\pm$0.21 &70.67$\pm$0.17 &57.17$\pm$0.61 &57.64$\pm$0.85 &56.98$\pm$0.88 &58.51$\pm$1.40 \\
\midrule
BiNE & 90.74$\pm$0.45 & 92.84$\pm$0.27 & 78.70$\pm$0.98 & 80.56$\pm$2.20 & 72.54$\pm$0.47 & 75.99$\pm$0.76 & 78.94$\pm$0.56 & 79.13$\pm$0.43 \\
BGNN-mlp & 76.78$\pm$2.07 & 83.05$\pm$1.59 & 68.32$\pm$0.23 & 74.11$\pm$0.23 & 61.96$\pm$0.64 & 65.22$\pm$0.67 & 66.54$\pm$0.23 & 68.84$\pm$0.18 \\
BGNN-adv & 90.35$\pm$1.80 & 92.35$\pm$1.11 & 83.47$\pm$0.16 & 84.70$\pm$0.15 & 77.49$\pm$1.10 & 77.26$\pm$1.01 & 82.76$\pm$0.09 & 82.40$\pm$0.10 \\
BiANE & 91.86$\pm$0.19 & 92.19$\pm$0.25 & 76.70$\pm$0.19 & 78.64$\pm$0.34 & 78.35$\pm$0.31 & 80.93$\pm$0.25 & 78.47$\pm$0.12 & 82.35$\pm$0.12 \\
\midrule
DualHGCN-sym & 93.53$\pm$0.28 & 94.54$\pm$0.21 & 85.47$\pm$0.69 & 87.98$\pm$0.61 & 86.86$\pm$0.41 & 88.50$\pm$0.44 & 84.53$\pm$0.55 & 86.72$\pm$0.44 \\
DualHGCN-asym & \textbf{93.85$\pm$0.25}$^{\ast}$ & \textbf{95.00$\pm$0.13}$^{\ast}$ & \textbf{86.69$\pm$0.26}$^{\ast}$ & \textbf{88.69$\pm$0.85}$^{\ast}$ & \textbf{87.57$\pm$0.41}$^{\ast}$ & \textbf{89.02$\pm$0.42}$^{\ast}$ & \textbf{85.54$\pm$0.80}$^{\ast}$ & \textbf{87.51$\pm$0.82}$^{\ast}$ \\
\bottomrule
\end{tabular}
\begin{tablenotes}
\item[$\ast$] Asterisks represent where DualHGCN's improvement over baselines is significant (one-sided rank-sum p-value <0.01).
\end{tablenotes}
\end{threeparttable}
\end{table*}

\begin{table}[htbp]
\caption{The AUROC and AUPRC values of DualHGCN and baselines on the task of link prediction. The initial features for Amazon and Alibaba are attributes.}\label{linkprediction2}
\centering
\begin{threeparttable}
\begin{tabular}{c|c|c|c|c}
\toprule
\multirow{2}{*}{METH} & \multicolumn{2}{c}{Amazon} & \multicolumn{2}{|c}{Alibaba}\\
\cline{2-5}
 & AUROC & AUPRC & AUROC & AUPRC \\
\midrule
Inits & 57.19$\pm$0.32 & 61.94$\pm$0.34 & 54.05$\pm$0.24 & 55.59$\pm$0.28 \\
\midrule
N2v &50.30$\pm$0.44 &55.44$\pm$0.47 &50.10$\pm$0.49 &51.52$\pm$0.29 \\
GSA &67.76$\pm$0.28 &70.10$\pm$0.22 &77.69$\pm$0.43 &76.79$\pm$0.49 \\
GCN &56.26$\pm$0.35 &58.19$\pm$0.41 &71.38$\pm$0.36 &69.16$\pm$0.18 \\
GAT &62.44$\pm$0.22 &67.11$\pm$0.13 &59.12$\pm$0.68 &59.73$\pm$0.66 \\
\midrule
HGNN & 77.41$\pm$0.20 & 81.14$\pm$0.13 & 63.16$\pm$0.27 & 66.16$\pm$0.20 \\
HCHA & 62.66$\pm$0.72 & 67.84$\pm$0.72 & 57.98$\pm$0.15 & 59.03$\pm$0.25 \\
MGCN & 64.52$\pm$1.10 & 71.34$\pm$0.67 & 50.54$\pm$0.25 & 52.43$\pm$0.98 \\
HGCN & 58.61$\pm$1.39 & 70.59$\pm$1.54 & 63.76$\pm$0.16 & 66.09$\pm$0.10 \\
\midrule
M2v++ &60.24$\pm$0.17 &63.58$\pm$0.21 &65.98$\pm$0.18 &70.97$\pm$0.20 \\
HAT &69.45$\pm$0.26 &72.22$\pm$0.20 &50.51$\pm$0.77 &52.09$\pm$0.95 \\
\midrule
BiNE & 78.70$\pm$0.98 & 80.56$\pm$2.20 & 78.94$\pm$0.56 & 79.13$\pm$0.43 \\
BGN-m & 68.97$\pm$1.54 & 73.80$\pm$1.11 & 67.62$\pm$0.34 & 70.92$\pm$0.21 \\
BGN-a & 79.43$\pm$0.33 & 81.48$\pm$0.36 & 83.42$\pm$0.25 & 81.53$\pm$0.23 \\
BiANE & 79.57$\pm$0.25 & 81.99$\pm$0.20 & 76.31$\pm$0.16 & 79.89$\pm$0.15 \\
\midrule
DHG-s & 84.87$\pm$0.75 & 87.19$\pm$0.65 & 86.27$\pm$0.56 & 88.07$\pm$0.45 \\
DHG-a & \textbf{86.55$\pm$0.15}$^{\ast}$ & \textbf{88.55$\pm$0.16}$^{\ast}$ & \textbf{86.76$\pm$0.46}$^{\ast}$ & \textbf{88.59$\pm$0.23}$^{\ast}$ \\
\bottomrule
\end{tabular}
\begin{tablenotes}
\tiny{\item[$\ast$] Asterisks indicate where improvement over baselines achieved by DualHGCN is significant (one-sided rank-sum p-value <0.01).}
\tiny{\item[$\dagger$] Some abbreviations are used in the table, `inits' short for `initial features', `N2v' short for `Node2vec', `GSA' short for `GraphSAGE', `HGCN' short for `HyperGCN', `M2v++' short for `Metapath2vec++', `BGN-m' short for `BGNN-mlp', `BGN-a' short for `BGNN-adv', `DHG-s' short for `DualHGCN-sym', and `DHG-a' short for `DualHGCN-asym'.}
\end{tablenotes}
\end{threeparttable}
\end{table}

\subsection{Results on Link Prediction (Q1)}
Table~\ref{linkprediction1} shows the results obtained by DualHGCN and baselines on all four datasets without attributes, and Table~\ref{linkprediction2} 
shows their performance on Amazon and Alibaba with attributes. 

Both Table~\ref{linkprediction1} and \ref{linkprediction2} show that DualHGCN significantly outperforms other baselines on both datasets without and with node attributes. In Table~\ref{linkprediction1}, DualHGCN-asym achieves the highest scores on all four datasets. For DTI, the AUROC and AUPRC score achieved by initial features, which trains the adjacency matrix with the tied autoencoder, are 63.43$\%$ and 72.34$\%$ respectively, and three bipartite network embedding methods achieve the similar highest metric score among baselines (BiNE, BGNN-adv, and BiANE) where the highest AUROC score achieved by BiANE (91.86$\%$) and the highest AUPRC score achieved by BiNE (92.84$\%$). However, DualHGCN-asym performs better than all baselines and achieves the highest score on the DTI dataset (93.85$\%$ for AUROC and 95.00$\%$ for AUPRC respectively). 
For Alibaba-s,  the AUROC and AUPRC score achieved by DualHGCN-asym are 87.57$\%$ and 89.02$\%$ respectively, which are both higher than the second highest scores achieved by BGNN-adv (78.35$\%$ for AUROC and 80.93$\%$ for AUPRC). Moreover, for Amazon and Alibaba with attributes, DualHGCN also achieves the highest metric scores (see Table~\ref{linkprediction2}). It demonstrates that our proposed DualHGCN method is effective both 
with and without the initial attribute information on nodes.

Comparing Table~\ref{linkprediction1} and \ref{linkprediction2} we see that training tied autoencoder with the adjacency matrix as initial features plays an 
important role in predicting unknown interactions. For instance, the AUROC and AUPRC scores achieved by initializing features with the adjacency matrix are both almost 15$\%$ higher than the scores achieved by just using the attributes as features. 
DualHGCN appears to be 
more robust to different initial features, compared to most other methods. 
We observe that performance gap for baseline methods, across the two tables, is large. In contrast, AUROC and AUPRC values of DualHDCN are similar across the two feature initializations.

\begin{table*}[htbp]
\centering
\caption{The micro-F1 and macro-F1 values of DualHGCN and baselines on the task of node classification ($\%$). Alibaba(adj) denotes the adjacency matrix is used as initial features, and Alibaba(attr) means attributes are preprocessed as initial features.}\label{nodeclass}
\begin{threeparttable}
\begin{tabular}{c|c|c|c|c|c|c}
\toprule
\multirow{2}{*}{Methods} & \multicolumn{2}{c}{Alibaba-s} & \multicolumn{2}{|c}{Alibaba(adj)} & \multicolumn{2}{|c}{Alibaba(attr)}\\
\cline{2-7}
 & micro-F1 & macro-F1 & micro-F1 & macro-F1 & micro-F1 & macro-F1 \\
\midrule
Initial features & 25.97$\pm$0.34 & 8.25$\pm$0.09 & 26.63$\pm$0.47 & 8.41$\pm$0.12 & 39.34$\pm$0.21 & 25.61$\pm$0.37 \\
\midrule
Node2vec &21.17$\pm$0.41 &20.09$\pm$0.30 &21.37$\pm$0.64 &19.95$\pm$0.43 &21.37$\pm$0.64 &19.95$\pm$0.43 \\
GraphSAGE &22.06$\pm$0.71 &19.40$\pm$0.62 &23.83$\pm$1.08 &20.00$\pm$0.72 &23.47$\pm$1.14 &19.52$\pm$0.43 \\
GCN &21.91$\pm$0.71 &19.51$\pm$0.38 &23.67$\pm$1.07 &19.32$\pm$0.33 &24.21$\pm$1.10 &18.33$\pm$0.93 \\
GAT &22.70$\pm$0.51 &19.70$\pm$0.48 &23.15$\pm$0.62 &19.98$\pm$0.68 &23.64$\pm$1.54 &18.70$\pm$0.34 \\
\midrule
HGNN & 25.59$\pm$0.97 & 9.06$\pm$0.93 & 26.77$\pm$1.03 & 13.03$\pm$1.84 & 32.82$\pm$0.74 & 21.51$\pm$0.75 \\
HCHA & 26.22$\pm$0.10 & 8.31$\pm$0.03 & 27.10$\pm$0.03 & 8.53$\pm$0.02 & 44.88$\pm$0.35 & 29.17$\pm$0.41 \\
MGCN & 25.93$\pm$1.00 & 19.51$\pm$0.92 & 26.49$\pm$0.53 & 11.36$\pm$1.37 & 40.48$\pm$1.51 & 30.47$\pm$1.91 \\
HyperGCN & 26.27$\pm$0.13 & 8.32$\pm$0.03 & 27.11$\pm$0.06 & 8.53$\pm$0.02 & 40.66$\pm$0.31 & 26.53$\pm$0.34 \\
\midrule
Metapath2vec++ &22.34$\pm$0.68 &20.32$\pm$0.60 &22.72$\pm$0.23 &20.13$\pm$0.46 &22.72$\pm$0.23 &20.13$\pm$0.46 \\
HAT &25.64$\pm$1.48 &15.06$\pm$1.57 &27.07$\pm$0.38 &16.14$\pm$0.42 &26.28$\pm$1.62 &16.05$\pm$1.33 \\
\midrule
BiNE & 26.65$\pm$0.93 & 20.77$\pm$0.98 & 26.98$\pm$0.44 & 20.64$\pm$0.57 & 26.98$\pm$0.44 & 20.64$\pm$0.57 \\
BGNN-mlp & 28.01$\pm$1.67 & 22.71$\pm$1.96 & 24.50$\pm$0.52 & 19.86$\pm$0.54 & 25.04$\pm$0.67 & 17.43$\pm$1.12 \\
BGNN-adv & 29.74$\pm$1.82 & 16.92$\pm$1.74 & 28.39$\pm$0.51 & 21.59$\pm$1.09 & 46.32$\pm$1.53 & 36.98$\pm$0.60 \\
BiANE & 22.29$\pm$0.45 & 19.91$\pm$0.54 & 22.65$\pm$0.20 & 20.01$\pm$0.53 & 22.84$\pm$0.63 & 19.89$\pm$0.47 \\
\midrule
DualHGCN-sym & 34.68$\pm$1.19 & 34.02$\pm$1.06 & 31.54$\pm$0.74 & 29.77$\pm$0.60 & 45.21$\pm$0.85 & 41.65$\pm$0.86 \\
DualHGCN-asym & \textbf{36.43$\pm$0.81}$^{\ast}$ & \textbf{35.73$\pm$1.20}$^{\ast}$ & \textbf{34.29$\pm$0.54}$^{\ast}$ & \textbf{33.95$\pm$0.36}$^{\ast}$ & \textbf{46.63$\pm$0.48} & \textbf{43.59$\pm$0.52}$^{\ast}$ \\
\bottomrule
\end{tabular}
\begin{tablenotes}
\item[$\ast$] Asterisks indicate significant improvement over baselines by DualHGCN (one-sided rank-sum p-value <0.01).
\end{tablenotes}
\end{threeparttable}
\end{table*}

\subsection{Results on Node Classification (Q2)}
Note that the Alibaba dataset contains attributes for each node, and we adopt two different ways to generate initial features. Alibaba(adj) denotes that we use the adjacency matrix as the initial features, and Alibaba(attr) means that attributes are used to generate initial features. 
For the smaller unattributed dataset Alibaba-s, only the adjacency matrix is used to generate initial features. 

The experimental results of both DualHGCN and baselines are summarized in Table~\ref{nodeclass}. The proposed DualHGCN performs significantly better than other baselines. The micro-F1 and macro-F1 of DualHGCN-asym achieved on Alibaba-s dataset are 36.43$\%$ and 35.73$\%$ respectively, which are significantly higher than the second highest score achieved by BGNN (29.74$\%$ for micro-F1 and 22.71$\%$ for macro-F1). 
The comparison between initial features and DualHGCN demonstrates the superior performance of DualHGCN on fusing the initial features and topological information to enhance the quality of unsupervised network embedding. 
Moreover, DualHGCN and baselines achieve higher metric scores on Alibaba (attr) than Alibaba (adj), which indicates that attributes of nodes contribute to improving the effects of node classification. For instance, the AUROC and AUPRC values achieved by DualHGCN-asym when initialized with the adjacency matrix as features are 34.29$\%$ and 33.95$\%$ respectively, which are about 10$\%$ lower than the scores achieved by the model with attributes as initial features. 

Note that the micro-F1 scores of many baselines are much larger than the corresponding macro-F1 scores, which indicates that these baselines result in classifications biased towards the large classes. For DualHGCN, the small gap between the micro-F1 and macro-F1 scores shows that DualHGCN is good at handling imbalanced classes for node classification.

\subsection{Effects of Message-passing Strategies (Q3)}
\begin{figure*}[!th]
\centering
\includegraphics[width=2.0\columnwidth]{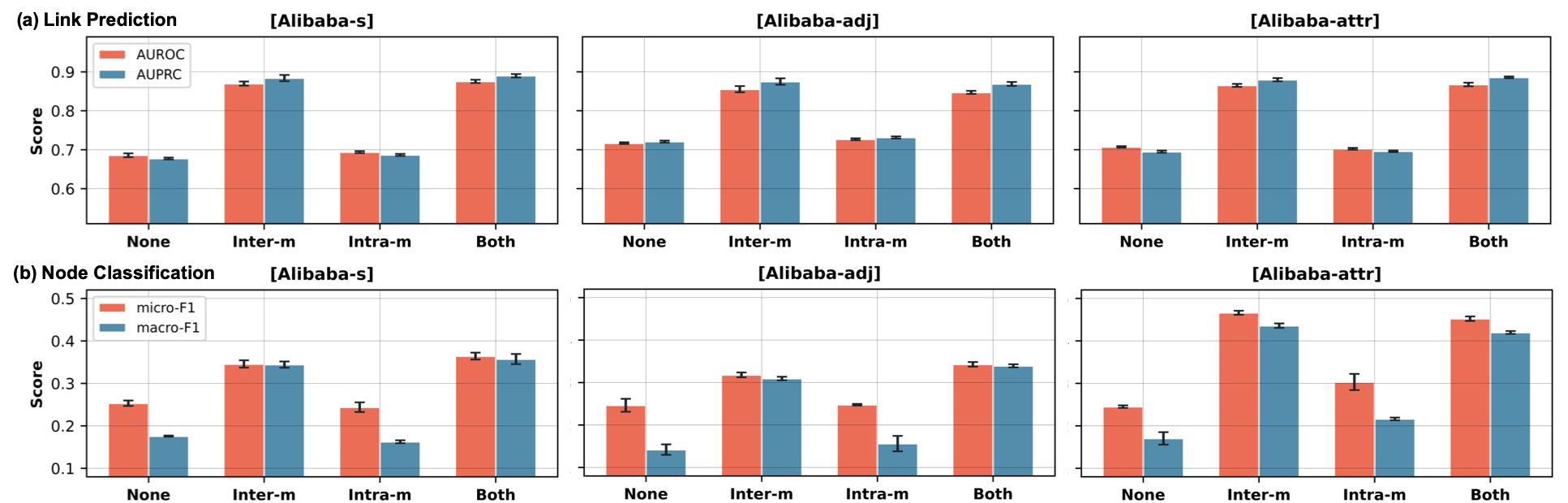}
\caption{The results of DualHGCN-asym with/without intra-/inter-message passing strategy on two datasets for link prediction and node classification tasks.} 
\label{fig3}
\end{figure*}

In DualHGCN, we propose two message passing strategies, intra-message passing, and inter-message passing. 
The intra-message passing strategy transfers learned features of `base' homo-hypergraph to other specific homo-hypergraphs, i.e., `click', `enquiry', and `contact' homo-hypergraphs, and the inter-message passing strategy shares features across dual homo-hypergraph sets (users and items) to enable communication between two distinct types of nodes. 
To answer \textbf{Q3}, we perform an ablation study on the DualHGCN model. Figure~\ref{fig3} shows the performance of DualHGCN-asym on Alibaba-s and Alibaba dataset with different message passing strategies. Overall, both intra- and inter-message passing strategies contribute to the tasks of link prediction and node classification, and the inter-message passing strategy plays an essential role in modeling real-world user behavior logs in the e-commerce platform. 

The effect of intra- and inter-message passing strategy depends on the distribution of nodes and edges among different types. In Alibaba, the inter-message passing strategy plays a more important role because of the imbalance of the average hyperedge degrees in different dual homo-hypergraphs. Note that the degree of a hyperedge is the number of nodes of the hypergraph incident to this hyperedge.

For instance, the average hyperedge degrees of different hypergraphs built from the Alibaba dataset are as follows, i.e., 2.28 for `base' homo-hypergraph, 2.23 for `click' homo-hypergraph, 1.79 for `enquiry' homo-hypergraph and 1.59 for `contact' homo-hypergraph respectively. 
Note that the gap of hyperedge degrees between `base' and `click'/`enquiry'/`contact' homo-hypergraphs is small, thus the contribution of the intra-message passing strategy on Alibaba is limited. The most conspicuous improvement of the intra-message passing is on Alibaba(attr) for the task of node classification. DualHGCN-asym without both intra- and inter- message passing strategy gets the 24.52$\%$ for micro-F1 and 17.04$\%$ for macro-F1 on Alibaba(attr) for the node classification task, and the micro-F1 and macro-F1 scores achieve 30.32$\%$ and 21.63$\%$ respectively if the intra-message passing is added into the DualHGCN-asym model. 


In contrast, the gap of hyperedge degrees between dual homo-hypergraph sets of users and items is large. For instance, in `base' homo-hypergraph, the average hyperedge degrees for users and items are 4.27 and 1.56 respectively. In this case, the effect of the inter-message passing strategy, which transfers information between two distinct domains (users and items), is essential. DualHGCN-asym without both intra- and inter- message passing strategy gets the 70.71$\%$ for AUROC and 69.50$\%$ for AUPRC on Alibaba(attr) for the link prediction task, and the AUROC and AUPRC scores achieve 86.53$\%$ and 87.99$\%$ when inter-message passing is added into the DualHGCN-asym model. 
In real-world datasets, especially in e-commerce, the imbalance of the average hyperedge degrees between users and items is common and difficult to model. Thus, the inter-message passing strategy plays an important role in modeling sparse dual homo-hypergraphs.

\subsection{Effects of Multiplex and Sparsity (Q4)}
To study the effect of multiplexing, we evaluate the performance of DualHGCN 
on each homo-hypergraph independently.
We also evaluate the performance of DualHGCN at different sparsity levels.

\begin{figure}[t]
\centering
\includegraphics[width=1.0\columnwidth]{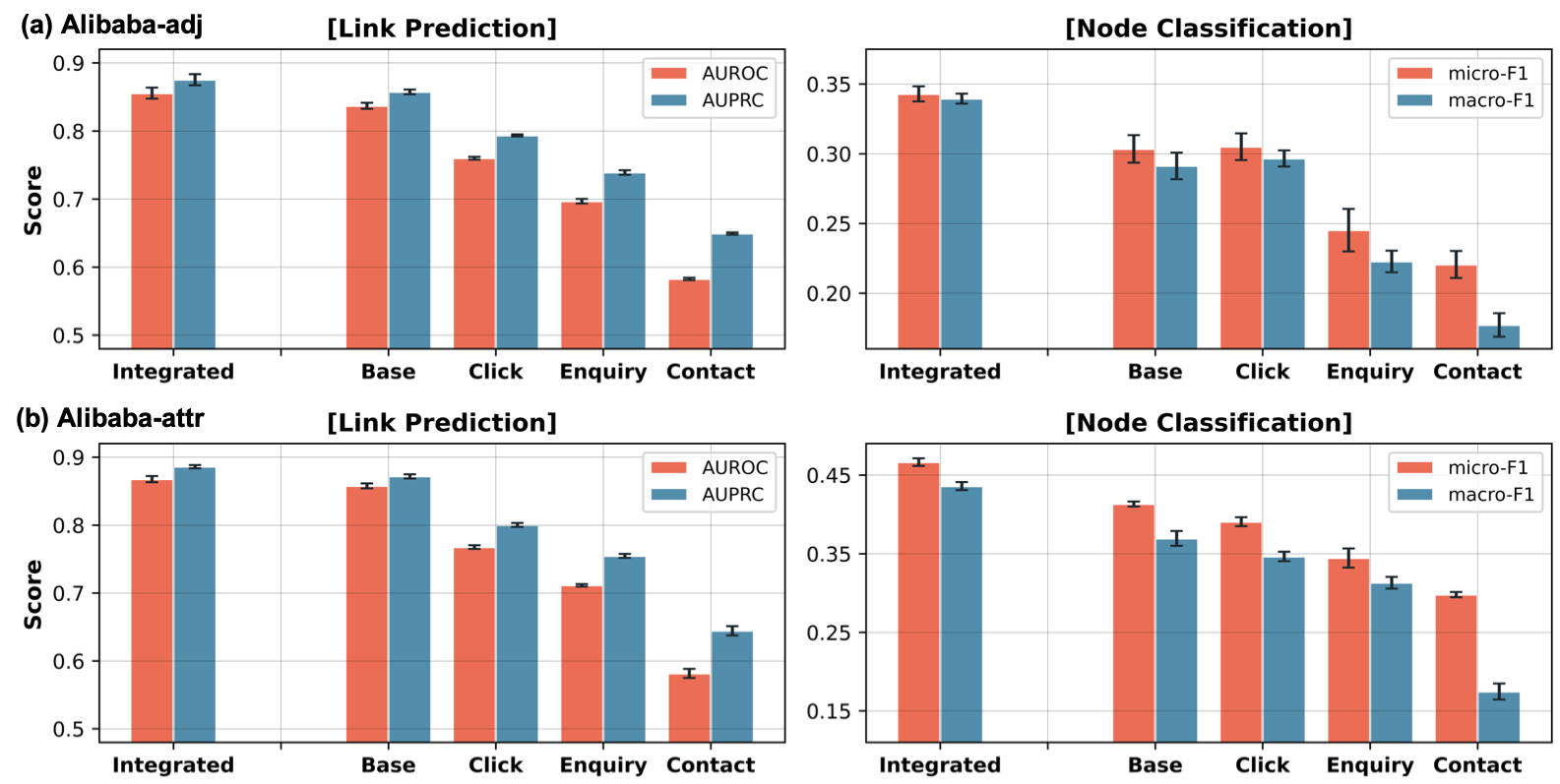}
\caption{The results of DualHGCN-asym on different edge-type datasets of Alibaba. The figure (a) and (b) use adjacency matrix and attributes as the initial features respectively.}
\label{fig4}
\end{figure}

\noindent\textbf{Effects of Multiple Edges.} 
We run the DualHGCN-asym method on each edge-type hypergraph (i.e., `base', `click', `enquiry' and `contact' homo-hypergraph) 
and compare it with DualHGCN-asym employed on all homo-hypergraphs. The experimental results are shown in Figure~\ref{fig4}. 
The results demonstrate the superior performance due to integrating different types of edges compared to treating these edges independently in each homogeneous hypergraph. For instance, the AUROC and AUPRC scores achieved by DualHGCN-asym on Alibaba(attr) for node classification are 46.63$\%$ and 43.59$\%$ respectively, which is significantly higher than other edge-type homo-hypergraphs (e.g., 41.31$\%$ for AUROC and 36.94$\%$ for AUPRC on `base' homo-hypergraph). 
Further, from Figure~\ref{fig4}, we find that the performance scores are correlated to the information enrichment of sub-bipartite network. 
We observe that the performance scores achieved on four homo-hypergraphs decrease progressively with decreasing number of edges in each sub-bipartite network 
(25,869 for `base', 25,180 for `click', 16,125 for `enquiry', 4,429 for `contact' sub-bipartite networks).

\noindent\textbf{Effects of Sparsity.}
We randomly delete a specific ratio of existing edges to increase the sparsity of the multiplex bipartite network, and employ DualHGCN, BiNE, BGNN-adv, and BiANE on these datasets to evaluate the performance. 

\begin{figure}[t]
\centering
\includegraphics[width=1.0\columnwidth]{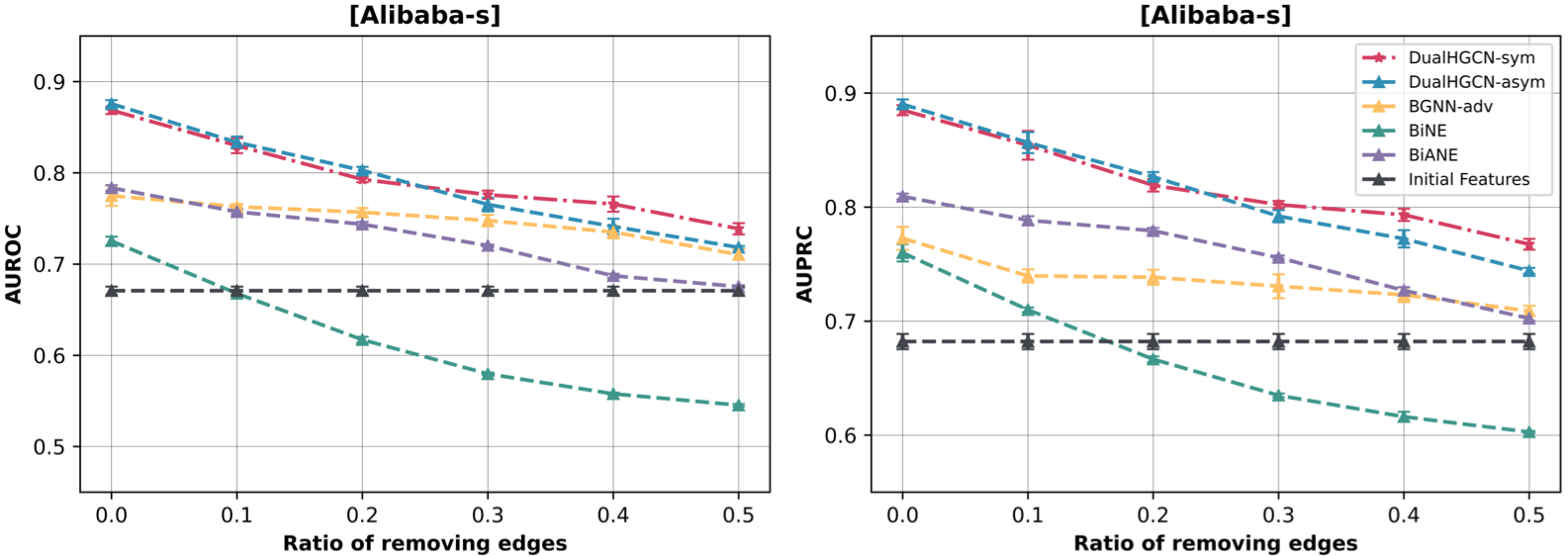}
\caption{The results of DualHGCN-sym/asym, BGNN-adv, BiNE, BiANE, and initial features on Alibaba-s with different sparsity of networks for link prediction.}
\label{fig5}
\end{figure}

Figure~\ref{fig5} shows that DualHGCN still significantly outperforms other baselines (BiNE, BGNN-adv and BiANE) when the networks become more sparse. 
With increase in sparsity, the performance of BiNE drops steeply compared to that of DualHGCN, BGNN-adv, and BiANE because BiNE focuses on the topological structure and does not utilize the initial features either from adjacency matrix or attributes. 
The rate of decrease in performance of DualHGCN is similar to that of BiANE. In extreme cases, when we randomly delete more than 50$\%$ edges, the performance scores achieved by DualHGCN is similar to the performance of initial features. Overall, DualHGCN has the best performance at various sparsity levels.

\subsection{Sensitivity Analysis and Visualization (Q5)}
We evaluate the sensitivity of DualHGCN to its three main (number of negative samples, number of layers, and parameter $\lambda$) and also qualitatively analyze the embeddings.

\begin{figure}[t]
\centering
\includegraphics[width=0.99\columnwidth]{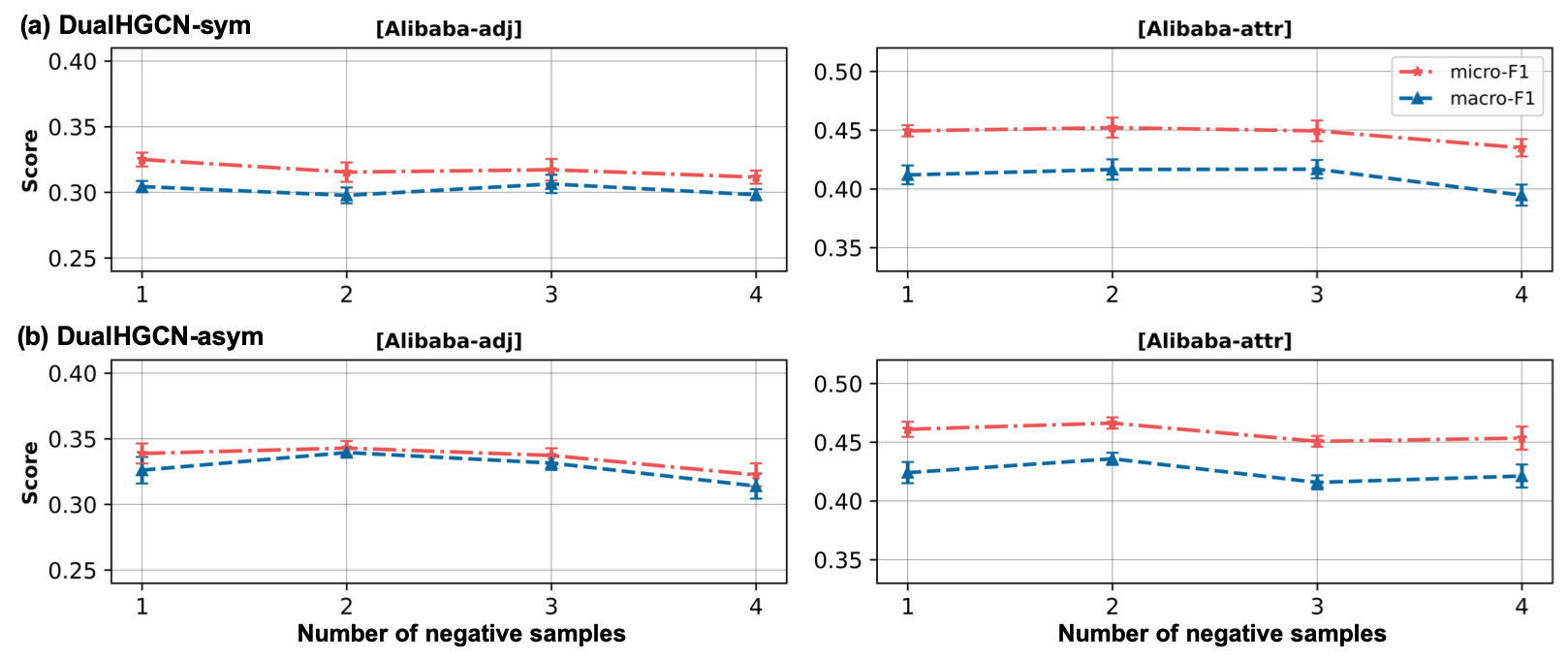}
\caption{The results of DualHGCN-sym/DualHGCN-asym on Alibaba with different numbers of negative samples for node classification.}
\label{fig6}
\end{figure}

\noindent\textbf{Effect of Negative Samples.}
To train our model, we randomly sample $n$ negative edges (unseen edges) for each positive edge (existing edge). 
The number of negative samples may affect the performance of final embedding. 
Here, we vary $n$ from 1 to 4, and evaluate the performance of DualHGCN on Alibaba-attr for the task of node classification (Figure~\ref{fig6}). Results demonstrate the robustness of our model DualHGCN on different number of negative samples. 

\begin{figure}[t]
\centering
\includegraphics[width=0.99\columnwidth]{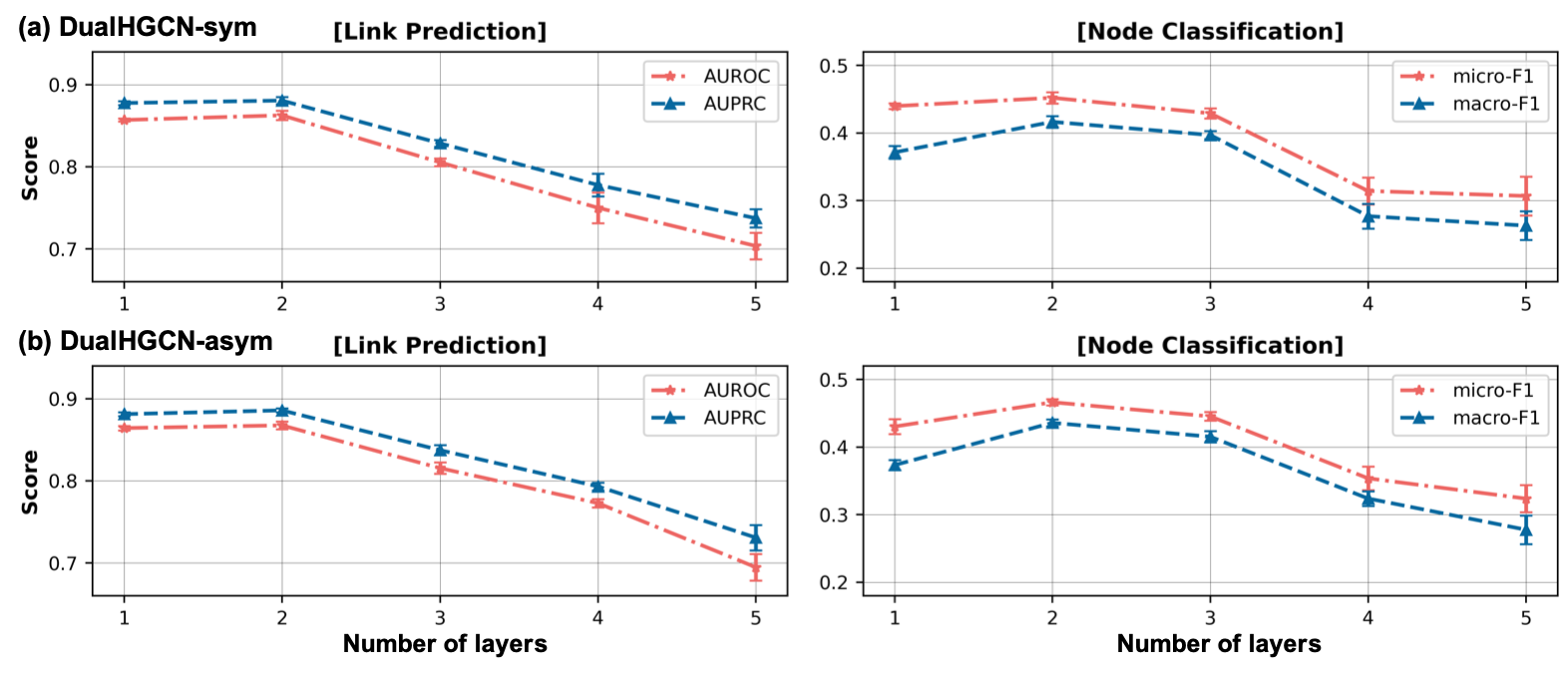}
\caption{DualHGCN-sym/asym on Alibaba with varying number of layers for link prediction and node classification.}
\label{fig7}
\end{figure}

\noindent\textbf{Effect of Layers.}
We investigate the performance of DualHGCN with different number of layers, ranging from 1 to 5. The results in Figure~\ref{fig7} show that DualHGCN achieves the best performance with two layers. With increase in number of layers, the performance of DualHGCN decreases slightly on both tasks of link prediction and node classification. This phenomenon has also been observed in classical graph convolutional networks~\cite{kipf2017gcn}. The reason stated in \cite{Li2018DeeperII} is that the graph convolutional operator is a special form of Laplacian smoothing. Increasing the number of layers makes it more difficult to train. Multiplication of Laplacian smoothing could lead to features of nodes being mixed and difficult to distinguish. This problem also exists in hypergraph convolutional operators.

\begin{figure}[t]
\centering
\includegraphics[width=0.99\columnwidth]{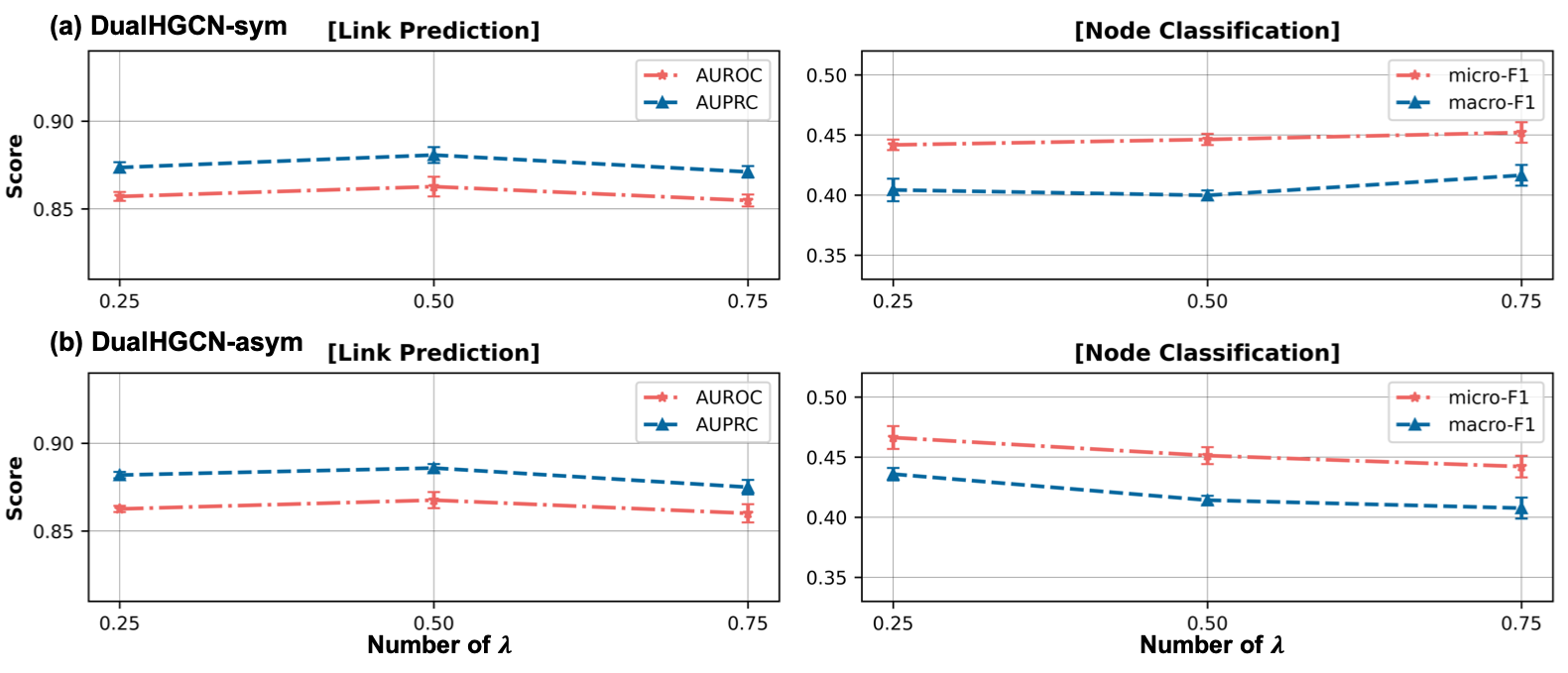}
\caption{DualHGCN-sym/asym on Alibaba(attr) with varying $\lambda$ for link prediction and node classification.}
\label{fig8}
\end{figure}

\noindent\textbf{Effect of Parameter $\lambda$.} 
The hyper-parameter $\lambda$ in the loss function is used to balance the importance between positive samples and negative samples. 
We investigate the effect of varying parameter $\lambda$, from 0.25 to 0.75, on both tasks of link prediction and node classification. 
From Figure~\ref{fig8}, we find that the performance of the model is not markedly sensitive to changes in $\lambda$ in both link prediction and node classification tasks. 

\begin{figure}[t]
\centering
\includegraphics[width=0.90\columnwidth]{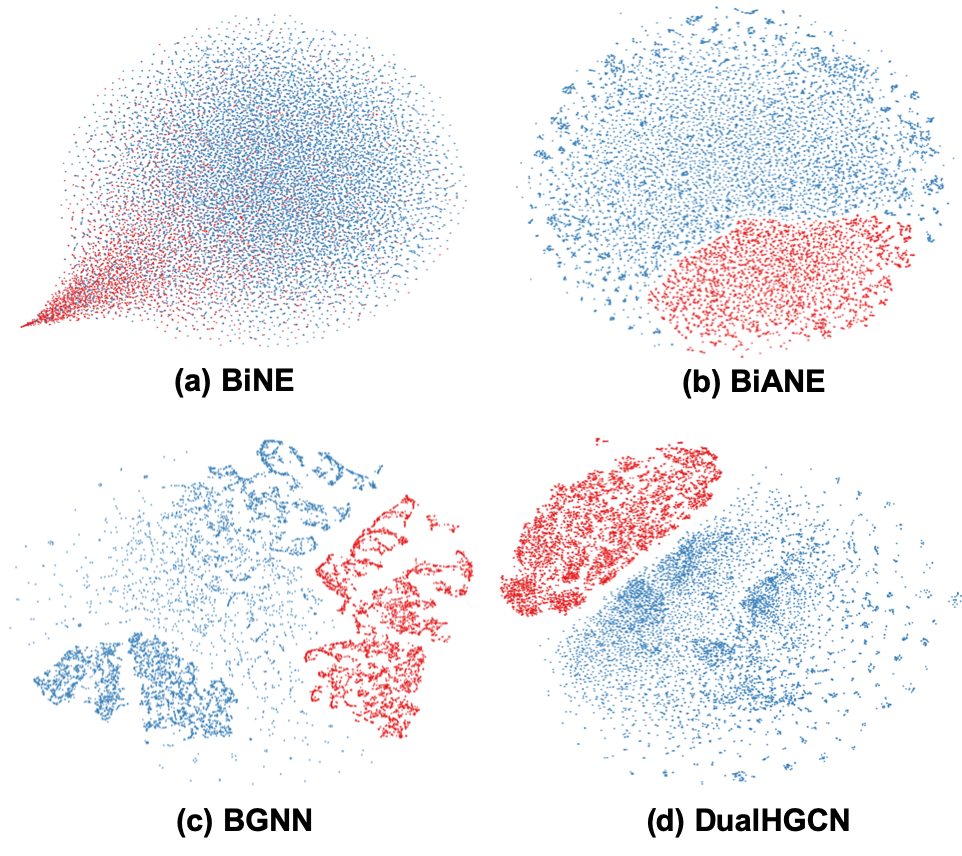}
\caption{Visualization of node embeddings on Alibaba with attributes dataset (red: users, blue: items).}
\label{fig11}
\end{figure}

\noindent\textbf{Visualization.} To conduct a qualitative assessment of the embeddings, we use the t-SNE~\cite{2008tsne} to visualize the final embeddings. Figure~\ref{fig11} shows the 2D-visualization of the embeddings from Alibaba network from DualHGCN-asym, BGNN-adv, BiANE and BiNE, where red nodes represent users and blue nodes represent items. 
BiNE produces embeddings in the same space for both users and items and thus visually the embeddings of the nodes are not well separated.
BiANE improves on BiNE in terms of separability because BiANE integrates the attribute information of two different types of nodes. 
BGNN also shows  good layout because it models the distinction between two types of nodes. Visually, DualHGCN gives the best separation between the two types of nodes. 

\section{Conclusion}
Multiplex bipartite networks appear in numerous important applications.
To our knowledge, our model DualHGCN is the first network embedding method that can model multiple edge types and node attributes in bipartite networks.
Further, it also effectively addresses common real-world challenges of sparsity and imbalance in node and edge type distributions.
The scalable transformation employed in DualHGCN to two sets of dual homogeneous hypergraphs enables the use of hypergraph convolutional operators on sparse inputs.
%
The intra-message passing strategy captures topological information across multiplex edges and addresses the problem of edge-type imbalance. The inter-message passing strategy tackles the challenges of node-degree and node-type imbalance between the two distinct node sets. 
Further, the DualHGCN architecture effectively uses node attributes when provided as inputs.
Our extensive experiments demonstrate the efficacy of DualHGCN on four real-world datasets for the tasks of link prediction and node classification. 
DualHGCN significantly outperforms 14 state-of-the-art methods from 4 different categories of embedding techniques.
They also highlight the strengths of our model with respect to robustness to varying sparsity levels, node attribute initialization strategies and handling of imbalanced classes.



\appendix

\section{Appendix}

\begin{figure}[t]
\centering
\includegraphics[width=0.85\columnwidth]{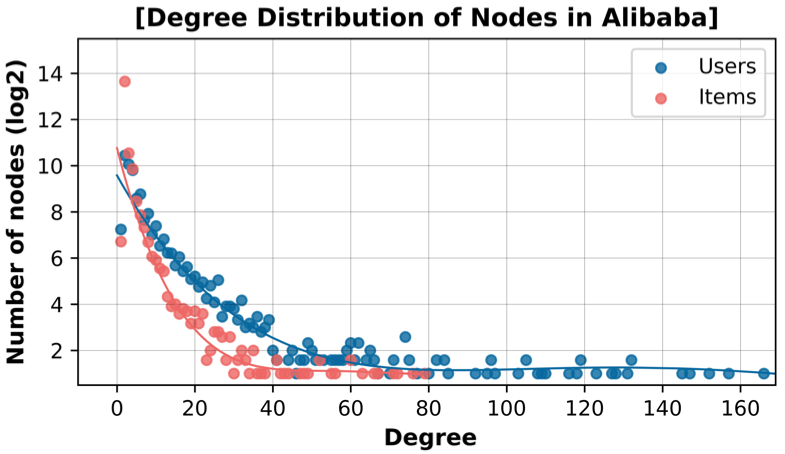}
\caption{Degree distribution of users, items in Alibaba.}
\label{fig9}
\end{figure}

\begin{figure}[t]
\centering
\includegraphics[width=0.85\columnwidth]{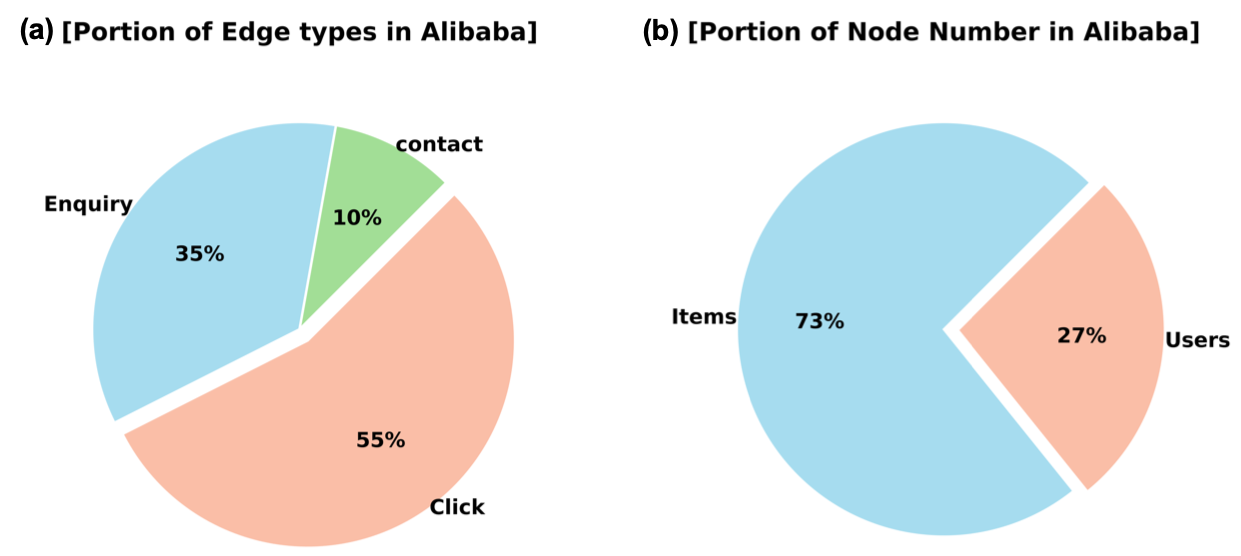}
\caption{Edge, Node type distributions in Alibaba dataset.}
\label{fig10}
\end{figure}

\textbf
{Supplementary Material}
Figure~\ref{fig9} shows the degree distribution of users and items in the Alibaba. 
Users have more rich and complicated structural information (e.g., users have more cases with degrees more than 2, and the degree of most of the items is 2.) 
Figure~\ref{fig10} shows the proportion of edge-type and node-number in the Alibaba. In Figure~\ref{fig10} (a), 55\% of edges is of type `click', which is more than other types of edges. In Figure~\ref{fig10} (b), the number of items are more than the number of users. These figures show the problem of edge-type and node-number imbalance in the data.

\textbf
{Implementation Details of DualHGCN}
In Table~\ref{parameters}, we show the parameters of DualHGCN-sym/-asym 
used in our experiments. 

\begin{table}[t]
\caption{Parameters of DualHGCN used in our experiments.}\label{parameters}
\centering
\begin{tabular}{c|c|c|c|c|c|c}
\toprule
\multirow{2}{*}{Parameters} & DTI & Amazon & \multicolumn{2}{|c}{Alibaba-s} & \multicolumn{2}{|c}{Alibaba}\\
\cline{2-7}
 & LP & LP & LP & NC & LP & NC \\
\midrule
lr & 0.002 & 0.002 & 0.001 & 0.001 & 0.002 & 0.005 \\\hline
Epochs & 4000 & 3000 & 3000 & 5000 & 3000 & 5000 \\\hline
Optimizer & \multicolumn{6}{c}{Adam} \\\hline
Dropout & 0.5 & 0.5 & 0.5 & 0.5 & 0.5 & 0.3 \\\hline
Weight decay & \multicolumn{6}{c}{5e-4} \\\hline
$\lambda$ & 0.5 & 0.5 & 0.5 & 0.5 & 0.75 & 0.25 \\\hline
Layers & \multicolumn{6}{c}{2} \\\hline
Neg samples & 2 & 3 & \multicolumn{2}{c|}{1} & 1 & 2 \\\hline
Inter & True & True & \multicolumn{2}{c|}{True} & True & True \\\hline
Intra & False & True  & \multicolumn{2}{c|}{True} & True & False \\\hline
Output emb & \multicolumn{5}{c}{32} \\
\bottomrule
\end{tabular}
\end{table}

\bibliographystyle{ACM-Reference-Format}
\balance
\bibliography{main}

\end{document}